\pdfoutput=1
\documentclass[journal]{IEEEtran}

\usepackage{cite}
\usepackage{amsmath,amssymb,amsfonts}
\usepackage[T1]{fontenc}
\usepackage{graphicx}
\graphicspath{{figures/}}
\usepackage{booktabs}
\usepackage{multirow}
\usepackage{tabularx}
\usepackage[flushleft]{threeparttable}
\usepackage{stfloats}
\usepackage{algorithm}
\usepackage{algorithmic}
\usepackage{subcaption}
\usepackage[table]{xcolor}
\usepackage{hyperref}
\usepackage{xspace}

\hypersetup{colorlinks=true,linkcolor=blue,citecolor=blue,urlcolor=blue}

\newcommand{\auspex}{\textsc{Auspex}\xspace}
\newcommand{\corvus}{\textsc{Corvus}\xspace}

\title{Fast Instinct, Slow Reason: A Dual-Speed Agentic Framework for Self-Configuring Multi-Object Tracking in Satellite Video}

\author{Fan~He,
        Baining~Wu,
        Yan~Li,~\IEEEmembership{Member,~IEEE,}
        and~Xiaoyang~Zeng,~\IEEEmembership{Senior~Member,~IEEE}%
\thanks{This work has been submitted to the IEEE for possible publication. Copyright may be transferred without notice, after which this version may no longer be accessible.}%
\thanks{This work was supported by the National Natural Science Foundation of China under Grant 62574049. (\textit{Corresponding author: Yan Li.})}%
\thanks{Fan He, Baining Wu, Yan Li, and Xiaoyang Zeng are with the State Key Laboratory of Integrated Chips and Systems, Fudan University, Shanghai 200433, China (e-mail: liyan@fudan.edu.cn).}%
}

\begin{document}

\maketitle

\begin{abstract}
Multi-object tracking in satellite video underpins persistent monitoring, yet offline-tuned trackers lack the onboard intelligence and efficiency to adapt within an observation of tens of seconds. The root is that adaptation still means relabelling and retraining the detector, which cannot run on board in time. Tuning the tracker configuration strongly controls accuracy without touching the network, yet search needs labelled or proxy evaluations, and a language agent costs seconds and tokens per decision. We propose a dual-speed agentic framework pairing a fast instinct compiled offline with a slow reason serving the mission. Its algorithm, Arbitration of Unlabelled Shortlists from Prior Experience, or \auspex, learns a covering shortlist and a label-free ranker from labelled source tasks and commits one configuration per sequence. Its system, a Cascaded Onboard Runtime with Verified Uplinked Skills, or \corvus, cascades reflex, skill, and agent layers, so that a verified skill commits 2.5 s into each observation while a language agent serves natural-language missions. On SatMTB and SatVideoDT with three recent detectors, \auspex raises IDF1 by 2.84 to 2.94 points and MOTA by 3.21 to 3.29 points on average without target labels, balances accuracy and labels best among sixteen methods, and loses at most 0.28 IDF1 points at 1.25 tracking passes. \corvus executes four missions with execution accuracy up to 0.99 and target recall at the perception ceiling, spends no token on the routine decision, runs on Jetson Xavier NX, and hosts fifteen language models from six families without altering any commitment, demonstrating its extensibility and generalization.
\end{abstract}


\IEEEpeerreviewmaketitle

\section{Introduction}
\label{sec:intro}

\begin{figure}[!t]
\centering
\includegraphics[width=\columnwidth]{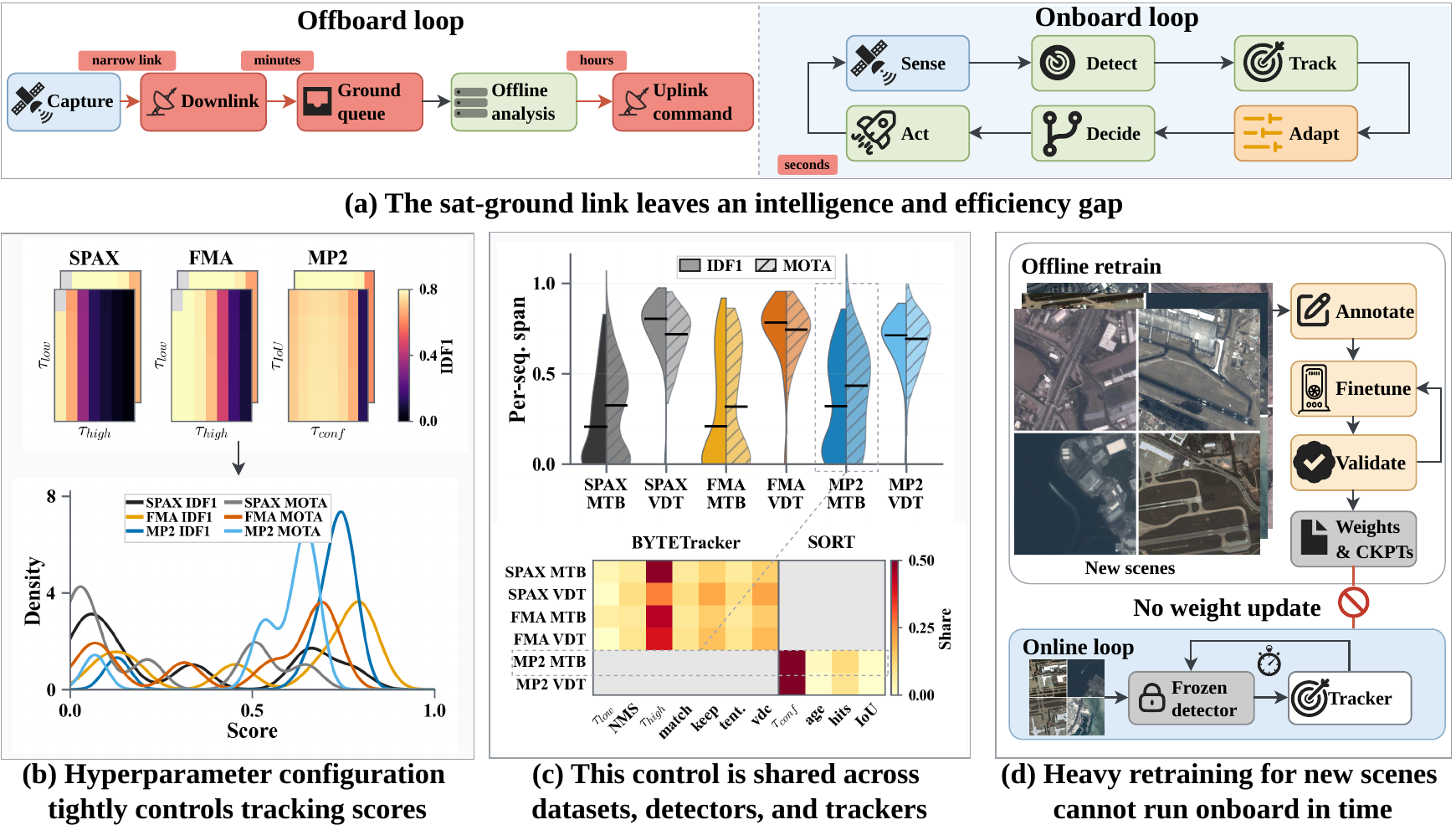}
\caption{Motivation for adapting the tracker configuration on board. (a) An offboard loop cannot act within one observation, (b) the configuration tightly controls tracking scores, (c) this control is shared across datasets, detectors, and trackers, and (d) retraining for new scenes cannot run on board in time.}
\label{fig:motivation}
\end{figure}

\IEEEPARstart{S}{atellite} video brings persistent observation to traffic monitoring, maritime surveillance, and airport operations, and multi-object tracking, abbreviated MOT, turns it into the trajectories of vehicles, ships, and aircraft~\cite{yin2022viso,li2023satmtb,satvideodt2022}. The targets span a few pixels, crowd among similar neighbours, and carry almost no appearance cue, and a line of trackers has raised accuracy on these benchmarks through better architectures~\cite{he2022tgram,xiao2022dsfnet,zhao2024mp2net,lu2025fmanet,xiao2026spax}. These accuracies, however, are measured offline with every setting tuned on data of the same kind, whereas an onboard closed loop asks for more. As Fig.~\ref{fig:motivation}a contrasts, a satellite must sense, detect, track, adapt, and act within an observation that lasts a median of 20 to 30 s, whereas an offboard loop of downlink, queueing, analysis, and uplink responds after minutes to hours. The gap is one of intelligence, because nothing on board adjusts the chain to the scene in view, and one of efficiency, because any adjustment made offboard arrives after the observation it was meant for.

What an onboard system can change without touching the network is the configuration of the post-detection stage, the thresholds and windows by which an association tracker such as SORT~\cite{bewley2016sort} or ByteTrack~\cite{zhang2022bytetrack} turns detections into tracks. This configuration is far from a detail. On the same frozen detections of SatVideoDT, the mean identity F1 score, abbreviated IDF1, that a configuration attains ranges from below 13 to about 80 percent for each of three recent detectors, and the median span over configurations on a single sequence is 71 to 80 percentage points, as Fig.~\ref{fig:motivation}b shows. The parameters that carry this control are shared across the two datasets, three detectors, and two trackers examined here, with the detection score threshold dominating every ByteTrack task in Fig.~\ref{fig:motivation}c. Changing the detector instead requires annotation, fine-tuning, and validation of new weights, a cycle that cannot run on board within an observation, as Fig.~\ref{fig:motivation}d illustrates. The configuration is therefore the lever an onboard system can pull, and the question becomes how to choose it for each sequence without labels.

Hyperparameter optimization treats the configuration as the argument of an unknown objective and searches it~\cite{bergstra2012random,snoek2012practical,jamieson2016non}, and its transfer variants carry experience over from labelled tasks~\cite{wistuba2015smfo,perrone2019boundingbox,salinas2020cts}, which suits a chain that must specialize to each domain yet cannot be retrained in time. Two properties nevertheless keep these methods off the satellite. Every step of a search ranks configurations by a score of what it has tried, which on board can only come from a proxy written by hand or extrapolated from other data~\cite{huang2020sqe,zhao2024hpod}, so the search follows the proxy wherever it departs from tracking accuracy, and the number of evaluations, and with it the moment a search settles, depends on the target, so neither its result nor its timing can be guaranteed within one observation.

Language agents, which couple a large language model with tools through interleaved reasoning and action~\cite{yao2023react,shen2023hugginggpt}, open another route to closing such loops, and remote sensing agents~\cite{xu2026rsagent,feng2026earthagent} and onboard language models~\cite{maranto2024llmsat,wu2026spacemind} carry them toward Earth observation and spacecraft control. For routine perception decisions, however, existing remote sensing agents run offline over static imagery and call perception models with default settings, every decision waits on a tool call that the model may or may not issue at a reasoning turn of seconds, a price a decision repeated on every sequence can hardly afford, and onboard studies report planning that degrades with mission complexity and latency misaligned with the process under control~\cite{maranto2024llmsat,mousist2025astrea,victoria2026navi,wu2026spacemind}. What is missing is a domain framework that keeps routine perception decisions fast, label-free, and verifiable on board while leaving goals and exceptions to the agent.

We observe that this routine decision needs neither a search nor a reasoning turn on board. The experience of labelled tasks can be compiled offline into a few complementary candidates and a learned judgment of their outputs, and only the final arbitration among them depends on the sequence in view. Building on this observation, we propose a framework that pairs an algorithm with a system. \auspex, for Arbitration of Unlabelled Shortlists from Prior Experience, compiles a covering shortlist and a label-free pairwise ranker from labelled source tasks, then on board commits one configuration per sequence from a short prefix. \corvus, a Cascaded Onboard Runtime with Verified Uplinked Skills, carries \auspex as a verified skill package and fires it on an event of the runtime, so that the language model above never stands on the path that commits the configuration. The framework thus runs at two speeds, a fast instinct of compiled skills that configures the tracker within milliseconds once the prefix ends, and a slow reason of the language model that interprets missions and exceptions over seconds, which makes the tracker self-configuring without making it wait for the agent. The contributions are twofold.
\begin{itemize}
\item We propose \auspex, which pairs a covering shortlist with a \(1-1/e\) coverage guarantee and a label-free pairwise ranker, and commits one configuration per sequence with no label or token. On the SatMTB and SatVideoDT benchmarks with three recent leading detectors and two trackers, it raises IDF1 by 2.84 to 2.94 points and the multiple object tracking accuracy, abbreviated MOTA, by 3.21 to 3.29 points on average without harming any task, and attains the best balance of accuracy and labels among sixteen methods. Committing from a 25-frame prefix cuts its cost from three to at most 1.25 tracking passes and its IDF1 gain by at most 0.28 points.
\item We design \corvus, which places each decision in the cheapest layer that meets its deadline, closes the configuration decision on board at the event-based autonomy level E3 of the European operability standard~\cite{ecss2025operability} 2.5 s into each observation, and uplinks skills 282 to 400 times smaller than detector weights. It executes missions stated in natural language with an execution accuracy of up to 0.99 and a target recall at the perception ceiling, without spending a language-model token on the routine decision. Fifteen language models from six families attach through one schema without altering any committed configuration, and new detectors, skills, and models can be admitted independently.
\end{itemize}

\section{Related Work}
\label{sec:related}

Choosing a tracker configuration on board without ground truth draws on four lines of work, each covering part of the problem, and the gaps between them shape both the selector and the runtime that carries it.

\subsection{Multi-Object Tracking in Satellite Video}
\label{sec:related-mot}

Satellite video trackers follow tracking by detection on targets of a few pixels, a regime made measurable by VISO~\cite{yin2022viso}, SatMTB~\cite{li2023satmtb}, and SatVideoDT~\cite{satvideodt2022}. Network methods such as TGraM~\cite{he2022tgram}, DSFNet~\cite{xiao2022dsfnet}, CFTracker~\cite{kong2023cftracker}, LocaLock~\cite{kong2025localock}, MP2Net~\cite{zhao2024mp2net}, FMA-Net~\cite{lu2025fmanet}, and SPAX~\cite{xiao2026spax} raise accuracy by learning detection and association jointly or by fusing motion cues across frames, yet they carry the association thresholds as constants and can follow a change of scene only through retraining on labelled imagery. Association trackers such as SORT~\cite{bewley2016sort}, ByteTrack~\cite{zhang2022bytetrack}, OC-SORT~\cite{cao2023ocsort}, and BoT-SORT~\cite{aharon2022botsort} link detections by motion and overlap, and adaptive variants such as AdapTrack~\cite{shim2024adaptrack}, the method of~\cite{ma2023adaptconf}, HRTracker~\cite{wu2024hrtracker}, and OITrack~\cite{lu2025oitrack} set one threshold from statistics of the current frame. Such a rule consults no experience from other scenes, so a threshold that suits one dataset can mislead on another, as Section~\ref{sec:exp-main} shows on SatVideoDT. Weight adaptation through CDTrack~\cite{zheng2023cdtrack}, DARTH~\cite{segu2023darth}, Tent~\cite{wang2021tent}, and the continual scheme of~\cite{yong2026ctaod} updates the tracker or detector on unlabelled target data, which needs gradients and a guard against drift and alters a payload that was qualified and launched frozen. \auspex instead keeps the detector frozen and selects among discrete configurations of the post-detection stage once per sequence.

\subsection{Hyperparameter Optimization with Labels}
\label{sec:related-hpo}

Hyperparameter optimization spends labelled evaluations to locate the best configuration. In tracking, Madrigal et al.~\cite{madrigal2018hpo}, Fleck and Zöllner~\cite{fleck2021bomot}, and MFGCS~\cite{adzemovic2026mfgcs} tune association parameters by model-based or coordinate search against annotated sequences, which confirms that the configuration matters but ties every decision to ground truth. Random search~\cite{bergstra2012random}, Gaussian-process Bayesian optimization, abbreviated GP-BO~\cite{snoek2012practical}, and successive halving~\cite{jamieson2016non} start from nothing, while ZeroShot~\cite{wistuba2015smfo}, BoundingBox~\cite{perrone2019boundingbox}, CTS~\cite{salinas2020cts}, and FSBO~\cite{wistuba2021fsbo} carry a portfolio, a search region, or a surrogate over from source tasks, and PFNs4BO~\cite{muller2023pfns4bo} and LLAMBO~\cite{liu2024llambo} bring a pretrained or a language-model prior. Transfer shortens the search but keeps its form, ranking configurations by labelled evaluations on the target until it converges online. The covering shortlist of Section~\ref{sec:method-shortlist} keeps from transfer the idea of complementary configurations but completes the search offline, so the target receives a few candidates instead of an optimizer.

\subsection{Label-Free Configuration Selection}
\label{sec:related-labelfree}

Label-free selection replaces the missing validation score with something observable on the target, in the tradition of algorithm selection~\cite{rice1976algorithm,xu2008satzilla}. Meta-feature predictors such as ISAC~\cite{kadioglu2010isac}, MetaOD~\cite{zhao2021metaod}, and ZAP-HPO~\cite{ozturk2022zap} map a description of the target data to a configuration without running the model on it, so an extrapolated choice for an unfamiliar dataset cannot be corrected. Output scores written by hand, such as SQE~\cite{huang2020sqe} and the track-count proxy of~\cite{davila2026vlmtracker}, observe the tracker but fix in advance what a good trajectory looks like, and on targets a few pixels wide they can reward permissive settings that add false tracks. Learned evaluators such as HPOD~\cite{zhao2024hpod} and HyPer~\cite{ding2024hyper} steer a search on the target with a proxy trained on labelled sources, which is credible only where the sources resemble the target, and a budget of a few passes leaves the search unfinished. Online selectors outside multi-object tracking, such as MEEM~\cite{zhang2014meem}, MCCT~\cite{wang2018mcct}, and Chameleon~\cite{jiang2018chameleon}, switch among experts frame by frame or take an expensive configuration as pseudo ground truth, and none faces the identity handover that a switch imposes on many targets at once. The arbiter of Section~\ref{sec:method-arbiter} instead judges only the few shortlisted candidates relative to one another and commits once per sequence after a short prefix.

\subsection{Language Agents and Onboard Autonomy}
\label{sec:related-agents}

Language agents couple a large language model with tools through interleaved reasoning and action~\cite{yao2023react,schick2023toolformer} and dispatch or compose vision models~\cite{shen2023hugginggpt,suris2023vipergpt}. Used as optimizers, as in OPRO~\cite{yang2024opro} and AgentHPO~\cite{liu2025agenthpo}, they pay for model inference at every decision, a cost that~\cite{kapoor2025agents} argues must be reported beside accuracy, and Voyager~\cite{wang2024voyager}, LATM~\cite{cai2024latm}, and AWM~\cite{wang2025awm} amortize it by distilling reasoning into skills, tools, or workflows whose invocation the agent still decides. Remote sensing agents such as RS-ChatGPT~\cite{guo2024rschatgpt}, Change-Agent~\cite{liu2024changeagent}, RS-Agent~\cite{xu2026rsagent}, GeoLLM-Engine~\cite{singh2024geollmengine}, and Earth-Agent~\cite{feng2026earthagent}, surveyed in~\cite{talemi2026agenticrs}, call interpretation models on the ground over static imagery with default settings, so the configuration of the perception chain is never the object of a decision.

On board, CASPER~\cite{knight2001casper} and the autonomous science software of EO-1~\cite{chien2005ase} replan observations from detected events, and the operability standard of the European Cooperation for Space Standardization, abbreviated ECSS~\cite{ecss2025operability}, grades such capability from execution under ground control to goal-oriented replanning, the levels E1 to E4. $\Phi$-Sat-1~\cite{giuffrida2022phisat} and CogniSAT-6~\cite{chien2025dynamic} run fixed networks in orbit, and the payload of~\cite{mateo2023worldfloods} retrains a model on the ground and uplinks its weights, so change arrives only as new weights. LLMSat~\cite{maranto2024llmsat}, ASTREA~\cite{mousist2025astrea}, NAVI-Orbital~\cite{victoria2026navi}, and SpaceMind~\cite{wu2026spacemind} place a language model in control of a spacecraft or its payload, and report that planning degrades with mission complexity or that inference latency misaligns with the process controlled. \corvus therefore compiles the routine configuration choice on the ground into a skill uplinked as data, fires it on board without a language model, and leaves a bounded agent with exceptions and goals, which limits the claim to event-based autonomy at level E3 for this decision.

\section{\auspex: Arbitration of Unlabelled Shortlists from Prior Experience}
\label{sec:method}

\begin{figure*}[t]
\centering
\includegraphics[width=\textwidth]{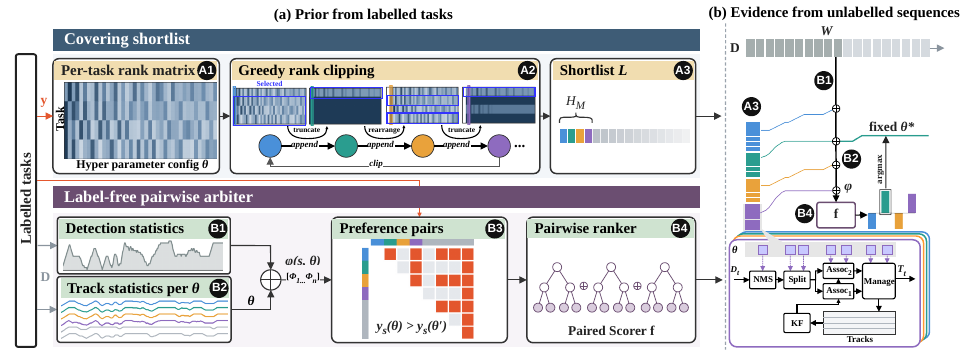}
\caption{Overview of \auspex. (a) Offline, labelled source tasks yield a covering shortlist and a pairwise ranker on label-free statistics. (b) On board, the head of the shortlist tracks a short prefix in parallel, and the ranker commits one configuration for the whole sequence.}
\label{fig:method}
\end{figure*}

The selectors of Section~\ref{sec:related} predict without observing the target, judge its output through a proxy written by hand or extrapolated from the sources, or spend labelled evaluations on an online search, and none settles on an unlabelled sequence at a cost known in advance. This section therefore presents \auspex, outlined in Fig.~\ref{fig:method}, which completes the search offline and leaves on board only a choice among a few candidates. A covering shortlist, learned once on labelled source tasks, decides which configurations are worth trying, and a label-free arbiter decides which of them to keep after a short prefix of each sequence, so that the tracker adapts to every sequence without labels, gradients, or language-model tokens at a cost fixed in advance, 1.17 to 1.25 tracking passes when it commits from a 25-frame prefix.

\subsection{Problem Formulation}
\label{sec:method-problem}

A frozen detector turns a sequence \(q\) of \(n_q\) frames into detections \(D_q=(D_q^{1},\dots,D_q^{n_q})\), and a tracker \(g\) links them into tracks \(\tau_q(\theta)=g(D_q;\theta)\) under a configuration \(\theta\) drawn from a finite grid \(\Theta\) of association parameters, the registers drawn in the inset of Fig.~\ref{fig:method}. The quality of the result, \(y_q(\theta)\), is its IDF1 against ground truth, which exists for the sequences of the labelled source tasks \(\mathcal{S}\) and never for a sequence observed on board. Only the post-detection stage is adapted, since it carries no learned weights and reruns cheaply, whereas the detector would need labels or gradients. \auspex fixes offline a head \(H_M\subseteq\Theta\) of \(M\) candidates and an arbiter \(a\), and commits on board
\begin{equation}
\theta^{*}_q = a\bigl(D_q^{1:W},\ \{\tau_q^{1:W}(\theta)\}_{\theta\in H_M}\bigr)
\label{eq:commit}
\end{equation}
from the first \(W\) frames alone, so that every frame of the sequence is tracked with \(\theta^{*}_q\). With \(\bar{y}_t(\theta)\) the mean quality of \(\theta\) over the sequences \(\mathcal{Q}_t\) of task \(t\), the regret \(R_t\) of the selector, the gap between \(\max_{\theta}\bar{y}_t(\theta)\) and the mean quality of the committed configurations, splits exactly into a term owned by the shortlist and a term owned by the arbiter,
\begin{align}
R_t &= R_{\mathrm{cov}}(M)+R_{\mathrm{arb}}(M), \nonumber\\
R_{\mathrm{cov}}(M) &= \max_{\theta\in\Theta}\bar{y}_t(\theta)-\max_{\theta\in H_M}\bar{y}_t(\theta), \label{eq:regret}\\
R_{\mathrm{arb}}(M) &= \max_{\theta\in H_M}\bar{y}_t(\theta)-\frac{1}{|\mathcal{Q}_t|}\sum_{q\in\mathcal{Q}_t}y_q(\theta^{*}_q). \nonumber
\end{align}
The coverage regret \(R_{\mathrm{cov}}\) cannot increase as the head grows, because heads are nested, while the arbitration regret \(R_{\mathrm{arb}}\) vanishes at \(M=1\) and tends to grow as more candidates compete, and it can turn negative when choices per sequence beat the best single member of the head. A good selector therefore needs a shortlist that drives \(R_{\mathrm{cov}}\) down at small \(M\) and an arbiter that keeps \(R_{\mathrm{arb}}\) small, and the two are designed separately below.

\subsection{Coverage-Based Shortlist Construction}
\label{sec:method-shortlist}

Absolute IDF1 differs widely between tasks, so the shortlist is built from ranks, following blocks A1 to A3 of Fig.~\ref{fig:method}. On each source task the rank of a configuration is
\begin{equation}
r_{t\theta} = 1 + \bigl|\{\theta'\in\Theta : \bar{y}_t(\theta') > \bar{y}_t(\theta)\}\bigr|,
\label{eq:rank}
\end{equation}
and the shortlist \(\mathcal{L}=(\theta_1,\theta_2,\dots)\) grows greedily by greedy rank clipping~\cite{wistuba2015smfo}. With \(c_t^{(k)}=\min_{j\le k} r_{t\theta_j}\) the best rank that task \(t\) attains from the first \(k\) entries, the next entry is
\begin{equation}
\theta_{k+1} = \arg\min_{\theta\in\Theta\setminus\mathcal{L}} \sum_{t\in\mathcal{S}} \min\bigl(r_{t\theta},\, c_t^{(k)}\bigr),
\label{eq:greedy}
\end{equation}
so clipping each row at its current best rank removes the credit of tasks already served and turns the choice toward tasks the list still neglects. Each step equivalently maximizes the marginal gain of the coverage
\begin{equation}
C(\mathcal{A}) = \sum_{t\in\mathcal{S}}\Bigl(|\Theta|+1-\min_{\theta\in\mathcal{A}} r_{t\theta}\Bigr),
\label{eq:coverage}
\end{equation}
a facility-location function that is monotone and submodular, so the greedy head satisfies \(C(H_M)\ge(1-1/e)\max_{|\mathcal{A}|=M}C(\mathcal{A})\)~\cite{nemhauser1978analysis}. Once every task holds its best configuration, \(C\) is maximal and carries no further information, and the remaining configurations are re-ranked on every task to start a new round whose entries back up the previous ones. A head wider than the first round already attains the maximum of \(C\), so the guarantee holds at every width, and the order of \(\mathcal{L}\) is the order of marginal coverage. The width \(M\) then acts as insurance against shifts that the source tasks do not reveal, since \(R_{\mathrm{cov}}\) on an unseen task cannot be estimated from the sources.

\subsection{Label-Free Pairwise Arbitration}
\label{sec:method-arbiter}

The arbiter judges a candidate by what can be measured without ground truth, following blocks B1 to B4 of Fig.~\ref{fig:method}. For a sequence \(s\) and a configuration \(\theta\), the feature vector
\begin{equation}
\varphi(s,\theta) = \bigl[\,d(D_s);\ u\bigl(\tau_s(\theta), D_s\bigr);\ \theta\,\bigr]
\label{eq:features}
\end{equation}
concatenates 24 detection statistics \(d\) of the scene, on counts above several score thresholds, score quantiles, target sizes, class shares, and their variation over time, with 16 track statistics \(u\) of the output, on identities per frame, track length and persistence, gap, jump, and fragmentation rates, motion, frame coverage, and the share of detections kept as tracks. A ranker \(f=\sum_{b=1}^{B}h_b\) of \(B\) regression trees is trained on every source sequence \(s\) with the pairwise logistic loss
\begin{equation}
\mathcal{J}(f)=\sum_{s}\sum_{(\theta,\theta')\in\mathcal{P}_s}\log\Bigl(1+e^{-[f(\varphi(s,\theta))-f(\varphi(s,\theta'))]}\Bigr),
\label{eq:pairwise}
\end{equation}
where \(\mathcal{P}_s\) holds the pairs with \(y_s(\theta)>y_s(\theta')\), so ground truth enters only through these pairs and the ranks of \eqref{eq:rank}, never through the features. Pairs are formed only within a sequence because crowded scenes lower every configuration together, so a ranker trained across sequences would learn which sequences are easy rather than which configurations are good, whereas \eqref{eq:pairwise} ignores any offset shared by one sequence and matches the decision on board among candidates on the same sequence. Which feature groups enter \(\varphi\) is chosen by leaving one source task out at a time. On a target sequence the scores of the \(M\) candidates are standardized as \(z_m=(f(\varphi_m)-\mu)/\sigma\), the committed configuration is the one with the largest \(z_m\), and the margin \(\delta=z_{(1)}-z_{(2)}\) between the two best is reported to the runtime of Section~\ref{sec:system}. Because every candidate already performs well on some source task, the arbiter only needs to order a few plausible options, a relative judgment that transfers across datasets more reliably than an absolute prediction of IDF1.

\subsection{Prefix-Based Online Commitment}
\label{sec:method-prefix}

Algorithm~\ref{alg:auspex} gives both halves of the method. On board, as the right panel of Fig.~\ref{fig:method} shows, the \(M\) candidates track the first \(W\) frames in parallel, the arbiter scores them from statistics of that prefix, and only \(\theta^{*}_q\) continues, reusing the tracks it has already produced, so the output of the whole sequence comes from one configuration and no identity is handed over. The decision is taken once per sequence because each later switch severs the identities that IDF1 rewards, and the prefix is short because the output is delayed by \(W\) frames, so \(W\) trades evidence against delay, from 2.5 s into the observation at \(W=25\) and ten frames per second to the whole sequence at \(W=n_q\).

\begin{algorithm}[t]
\caption{\auspex Compilation and Commitment}
\label{alg:auspex}
\footnotesize
\begin{algorithmic}[1]
\REQUIRE labelled sweeps \(y_s(\theta)\) on source tasks \(\mathcal{S}\), list length \(K\)
\STATE compute ranks \(r_{t\theta}\) by \eqref{eq:rank}, set \(\mathcal{L}\gets\emptyset\) and \(c_t\gets|\Theta|+1\)
\FOR{\(k=1\) \TO \(K\)}
  \IF{\(c_t=1\) for every \(t\in\mathcal{S}\)}
    \STATE re-rank \(\Theta\setminus\mathcal{L}\) on every task and reset \(c_t\)
  \ENDIF
  \STATE append \(\theta_k\) of \eqref{eq:greedy} to \(\mathcal{L}\), set \(c_t\gets\min(c_t,r_{t\theta_k})\)
\ENDFOR
\STATE fit \(f\) by minimizing \eqref{eq:pairwise}, \textbf{return} skill \((\mathcal{L},f)\)
\STATE \textbf{on board, for each sequence} \(q\) \textbf{with head} \(H_M=(\theta_1,\dots,\theta_M)\)
\FOR{\(i=1\) \TO \(W\)}
  \STATE advance one tracker per \(\theta_m\in H_M\) on \(D_q^{i}\)
\ENDFOR
\STATE \(z\gets\) standardized \(f(\varphi(q,\theta_m))\) over \(m\), \(\theta^{*}_q\gets\theta_{\arg\max z}\), \(\delta\gets z_{(1)}-z_{(2)}\)
\STATE stop the other trackers, keep \(\tau_q^{1:W}(\theta^{*}_q)\)
\FOR{\(i=W+1\) \TO \(n_q\)}
  \STATE advance the tracker of \(\theta^{*}_q\) on \(D_q^{i}\)
\ENDFOR
\RETURN \(\tau_q(\theta^{*}_q)\) and \(\delta\)
\end{algorithmic}
\end{algorithm}

\subsection{Complexity and Cost Analysis}
\label{sec:method-cost}

All costs that scale with the configuration space fall on the ground. The labelled sweep runs \(\sum_{t\in\mathcal{S}}|\mathcal{Q}_t||\Theta|\) tracking passes once, a step any transfer method shares, the greedy list costs \(O(K|\mathcal{S}||\Theta|)\) for \(K\) entries, and fitting \(f\) grows linearly with the number of trees, features, and source rows up to a logarithmic factor, none of which reaches the satellite. On board, a sequence costs a number of tracker frame updates
\begin{equation}
\kappa_q = MW + (n_q - W) = n_q\Bigl(1+\frac{(M-1)\,W}{n_q}\Bigr),
\label{eq:cost}
\end{equation}
which with \(M=3\), \(W=25\), and sequences of 200 to 300 frames equals 1.17 to 1.25 passes, against \(M\) passes when the whole sequence is scored with \(W=n_q\). The statistics in \eqref{eq:features} grow linearly with the boxes of the prefix and the ranker makes at most 5,400 comparisons for \(M=3\) and \(B=300\) trees of depth six, so the decision needs no label, gradient, or token and arrives at a fixed frame \(W\) whatever the target. A search with \(n\) evaluations instead runs \(n\) passes and consumes \(n\) labelled or proxy-scored evaluations, a Gaussian-process surrogate adds \(O(n^{3})\) per refit, and a language-model optimizer whose prompt carries the history of trials consumes tokens that grow quadratically with \(n\), while the time at which any of them settles depends on the target.

\section{\corvus: Cascaded Onboard Runtime with Verified Uplinked Skills}
\label{sec:system}

The selector of Section~\ref{sec:method} needs on board only a prefix, a ranker, and a few tracker instances, but does not itself decide when it runs, what it may change, how the ground updates it, or how it coexists with a language agent, while the onboard language models and fixed networks of Section~\ref{sec:related-agents} lag behind the process they control or change only through new weights. This section therefore presents \corvus, which cascades decisions across a reflex, a skill, and an agent layer by time scale, uplinks \auspex as a verified skill package instead of as weights, and keeps the language model off the path that commits the configuration. The reflex and skill layers form the fast path, acting within milliseconds, and the agent layer the slow path, reasoning over seconds, which realizes the dual-speed principle of Section~\ref{sec:intro}, so the configuration takes effect 2.5 s into every observation whether or not a language model responds while the agent retains goals and exceptions and the link carries kilobytes per update.

\subsection{Requirements and Decision Placement}
\label{sec:sys-req}

A satellite video observation in the two benchmarks lasts a median of 20 to 30 s, or 200 to 300 frames at ten frames per second, so an adaptation is worth having only if it takes effect within that window. Of five requirements, R1 asks the configuration to take effect within the observation, R2 to use no labels on board, R3 to leave the detector weights untouched, R4 to stay within the compute and power of an onboard processor, and R5 to accept mission goals that operators state in natural language. Every decision \(d\) is then assigned to the cheapest layer that meets its deadline \(T_d\) and these requirements,
\begin{align}
\ell^{*}(d) &= \arg\min_{\ell\in\Lambda(d)} c_\ell, \label{eq:placement}\\
\Lambda(d) &= \{\ell : \lambda_\ell\le T_d,\ \ell \text{ meets R2 to R4}\}, \nonumber
\end{align}
where \(\lambda_\ell\) and \(c_\ell\) are the latency and cost of layer \(\ell\). Four layers span the time scales of the mission, the reflex layer tracking every frame with the configuration it is given, pre-planned execution at ECSS level E2~\cite{ecss2025operability}, the skill layer committing a configuration at the prefix of each sequence, event-based autonomy at E3, the agent layer interpreting goals and exceptions over seconds to minutes toward E4, and the ground layer compiling skills between passes at E1. The configuration has a deadline of one prefix and needs no language, so \eqref{eq:placement} places it in the skill layer.

\begin{figure*}[t]
\centering
\includegraphics[width=\textwidth]{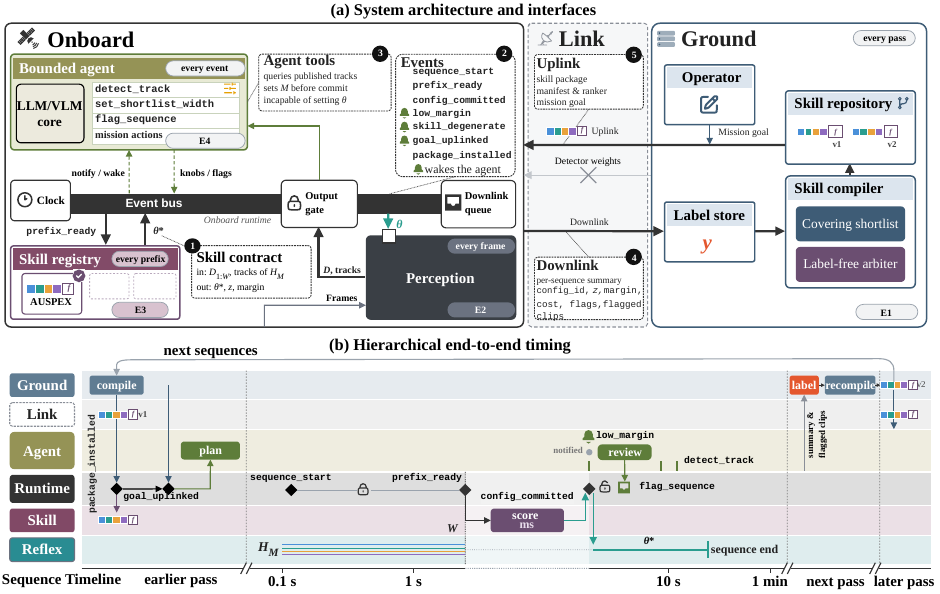}
\caption{Overview of \corvus. (a) Onboard, link, and ground components, where the routine path commits \(\theta\) through the skill registry without passing the agent. (b) Timing of one sequence across the layers on a logarithmic axis, with ground segments not to scale and agent latencies from one measured run.}
\label{fig:system}
\end{figure*}

\subsection{System Architecture and Interfaces}
\label{sec:sys-arch}

The upper panel of Fig.~\ref{fig:system} separates components by what they hold. Perception holds the frozen detector and tracker, whose only writable port is the configuration register \(\theta\), the runtime holds the clock, the event bus, an output gate, and the downlink queue but decides nothing itself, the skill registry holds installed packages such as the \texttt{tracker\_config} skill that carries the online half of \auspex, and the bounded agent holds a language model and its tools. Of the five numbered interfaces, the skill contract maps the detections \(D_q^{1:W}\) and the prefix tracks of \(H_M\) to \(\theta^{*}_q\), the standardized scores, and the margin \(\delta\) without labels or language-model calls, the event interface wakes the agent only on a low margin, a degenerate skill, or a new goal, and the agent tools, declared in one schema so that a model is attached by name, query tracks, set \(M\), flag sequences, and act on missions. No tool sets \(\theta\), so the routine path, drawn bold, runs from perception through the bus to the registry and back to \(\theta\) without passing the agent, which reads tracks only behind the output gate.

\subsection{Runtime Workflow}
\label{sec:sys-runtime}

Algorithm~\ref{alg:runtime} gives the workflow of one sequence, which the lower panel of Fig.~\ref{fig:system} traces. The runtime, not the agent, owns the clock and fires the skill on \texttt{prefix\_ready}, so the commitment never waits for a language model and never depends on a model choosing to call a tool. The output gate stays closed during the prefix and opens at commitment with the prefix tracks of \(\theta^{*}_q\), so every track the agent reads comes from the committed configuration and no identity it has seen is later withdrawn. A margin below the threshold \(\tau_\delta\) or an empty output from every candidate wakes the agent, which may flag the sequence for labelling on the ground and so close the slow loop, but cannot overrule the commitment.

\begin{algorithm}[t]
\caption{Runtime Workflow of \corvus for One Sequence}
\label{alg:runtime}
\footnotesize
\begin{algorithmic}[1]
\REQUIRE installed skill \((\mathcal{L},f,W,\tau_\delta)\), width \(M\) possibly set by the agent
\STATE on \texttt{sequence\_start}, start trackers for \(H_M\) and close the output gate
\FOR{each frame \(i=1,\dots,n_q\)}
  \STATE run the detector and every active tracker on frame \(i\)
  \IF{\(i=W\)}
    \STATE emit \texttt{prefix\_ready}, obtain \((\theta^{*}_q,z,\delta)\) from the skill
    \STATE write \(\theta^{*}_q\) to the register, stop the other trackers
    \STATE open the gate, publish \(\tau_q^{1:W}(\theta^{*}_q)\), emit \texttt{config\_committed}
    \IF{\(\delta<\tau_\delta\) \OR every candidate output is empty}
      \STATE emit \texttt{low\_margin} or \texttt{skill\_degenerate} to wake the agent
    \ENDIF
  \ELSIF{\(i>W\)}
    \STATE publish the tracks of \(\theta^{*}_q\) for frame \(i\)
  \ENDIF
\ENDFOR
\STATE queue the summary \((\theta^{*}_q,z,\delta,\kappa_q,\text{events},\text{flags})\) for downlink, with the clip if flagged
\end{algorithmic}
\end{algorithm}

\subsection{Onboard Computational Load Analysis}
\label{sec:sys-compute}

The three onboard layers run per frame, per sequence, and per event, and the processing time of one sequence of \(n_q\) frames decomposes accordingly,
\begin{equation}
T_q = \underbrace{n_q c_{\mathrm{det}} + \kappa_q c_{\mathrm{trk}}}_{\text{reflex}}
    + \underbrace{M c_{\varphi} + c_f}_{\text{skill}}
    + \underbrace{k\,\lambda_{\mathrm{A}}}_{\text{agent}},
\label{eq:load}
\end{equation}
where \(c_{\mathrm{det}}\) and \(c_{\mathrm{trk}}\) are the per-frame times of the detector and of one tracker, \(c_{\varphi}\) the time to compute the statistics of one candidate, \(c_f\) the time of the ranker, \(k\) the number of language-model turns the sequence triggers, and \(\lambda_{\mathrm{A}}\) the latency of one turn. The reflex term is the only one with a hard deadline on every frame,
\begin{equation}
c_{\mathrm{det}} + m_i\,c_{\mathrm{trk}} \le \frac{1}{\nu}, \qquad m_i=\begin{cases}M, & i\le W,\\ 1, & i>W,\end{cases}
\label{eq:realtime}
\end{equation}
for frame rate \(\nu\), so the prefix is feasible whenever the \(M-1\) extra trackers fit into the slack of the steady state, and only \(W\) frames carry them, which gives the \(\kappa_q\) of \eqref{eq:cost}. Once per sequence, the skill costs \(c_{\varphi}=O(W\bar{N})\) for one sweep over the \(\bar{N}\) boxes per frame of the prefix and \(c_f=O(MB)\) for the ranker of Section~\ref{sec:method-cost}, a single burst whose share per frame, \((Mc_{\varphi}+c_f)/n_q\), shrinks as sequences lengthen. The agent term follows events rather than frames, with a duty cycle
\begin{equation}
u_{\mathrm{A}} = \frac{k\,\lambda_{\mathrm{A}}\,\nu}{n_q}
\label{eq:duty}
\end{equation}
that is zero on the routine path and grows only with exceptions and mission actions. Because a language-model turn on an onboard processor lasts seconds, a commitment routed through the agent would release tracks no earlier than \(W/\nu+\lambda_{\mathrm{A}}\), at least twice the delay of the skill whenever \(\lambda_{\mathrm{A}}\ge W/\nu\), the latency argument behind \eqref{eq:placement}. Load therefore concentrates where it is cheapest, with perception continuous but light, the skill a burst once per sequence, and the language model idle except on events, a burst-and-wait pattern that a processor can exploit by keeping the model resident and asleep.

\subsection{Space-Ground Communication Load Analysis}
\label{sec:sys-link}

The link is the scarcest resource of an observation satellite, and \corvus sends across it in each direction only what the other side lacks. On the downlink, every sequence yields the summary of Algorithm~\ref{alg:runtime}, a few hundred bytes or less than one part in 5,000 of one uncompressed 1080 by 1080 colour frame, and a clip only when flagged. For \(N\) sequences of which a fraction \(p_{\mathrm{f}}\) is flagged, the downlink volume
\begin{equation}
B_{\mathrm{down}} = N\bigl(s_{\mathrm{sum}} + p_{\mathrm{f}}\,s_{\mathrm{clip}}\bigr)
\label{eq:downlink}
\end{equation}
is governed by \(p_{\mathrm{f}}\) and not by the frame rate or the duration of observation. On the uplink, the ground compiles Algorithm~\ref{alg:auspex} into a package of the ranker and a manifest recording the head of \(\mathcal{L}\), the feature columns, \(W\), \(M\), \(\tau_\delta\), the source tasks, and a hash of the ranker, which the registry verifies on installation, refusing a package whose sources include the task it is to serve. A package occupies 244 to 332 KiB after compression against 91 to 112 MiB for the weights of the three detectors, so a skill transfers 282 to 400 times faster than the detector it serves and is sent only when the detector or the imagery changes. More decisive than volume is when an update takes effect, since any ground loop must downlink the evidence, label it, and uplink the result,
\begin{equation}
\lambda_{\mathrm{G}} \ \ge\ t_{\mathrm{down}}+t_{\mathrm{label}}+t_{\mathrm{up}} \ \gg\ \frac{n_q}{\nu},
\label{eq:groundloop}
\end{equation}
so a ground loop cannot improve the observation that revealed the need, whereas the skill commits after \(W/\nu=2.5\) s, within the first 8 to 13 percent of the observation. The ground therefore updates what the satellite knows, and the satellite applies it to every sequence without waiting for the ground.

\section{Experiments}
\label{sec:exp}

The experiments separate the gain of choosing a configuration on each sequence from the gain of the layer that commits it, comparing \auspex with meta-learning, label-free selection, and hyperparameter optimization, dismantling the selector, altering the system around it, and carrying the runtime onto an embedded board.

\subsection{Experimental Setup}
\label{sec:exp-setup}

\subsubsection{Configuration Spaces}
Two association backbones define the spaces \(\Theta\) of Section~\ref{sec:method-problem}. ByteTrack~\cite{zhang2022bytetrack} associates high-score and low-score detections in two stages over a grid of 178 configurations of seven parameters, the two detection thresholds, two overlaps, the retention of lost tracks, the tentative period of new tracks, and the weight on direction consistency. SORT~\cite{bewley2016sort}, the association baseline of the SatVideoDT challenge~\cite{satvideodt2022}, associates in one stage over a grid of 250 configurations of the confidence threshold, the maximum track age, the minimum number of hits, and the matching overlap. Each tracker also carries one fixed deployment configuration, tuned once on a single scenario and applied to every sequence like a setting uploaded before launch, which is the reference for every gain.

\subsubsection{Detectors, Datasets, and Tasks}
Detections come from three recent satellite video detectors, SPAX~\cite{xiao2026spax}, FMA-Net~\cite{lu2025fmanet}, and MP2Net~\cite{zhao2024mp2net}, which stay frozen throughout and are chosen because stronger detections leave the post-detection stage less to repair. SPAX and FMA-Net feed the ByteTrack implementation released with SPAX, and MP2Net, which couples detection and tracking end to end, has its association replaced by SORT so that its configuration space becomes explicit. The imagery comes from SatMTB~\cite{li2023satmtb} and SatVideoDT~\cite{satvideodt2022}, both at ten frames per second. A task pairs one detector with one split of one dataset, and the seven tasks, the three detectors on the training split of each dataset and SPAX on the SatMTB test split, hold 70 to 134 sequences each and 682 in total, so that tasks differ in detector, in dataset, or in both.

\subsubsection{Transfer Protocols}
Let \(\mathcal{T}\) denote the seven tasks, and let task \(t\) inherit the tracker \(g_t\) of its detector and the dataset \(D_t\) of its split. When \(t\) is held out, leave one task out, abbreviated LOTO, trains on every other task of the same tracker, and leave one dataset out, abbreviated LODO, further discards every source that shares the held-out dataset,
\begin{align}
\mathcal{S}^{\mathrm{LOTO}}_{t} &= \{\, t' \in \mathcal{T}\setminus\{t\} \mid g_{t'} = g_{t} \,\}, \label{eq:loto}\\
\mathcal{S}^{\mathrm{LODO}}_{t} &= \{\, t' \in \mathcal{S}^{\mathrm{LOTO}}_{t} \mid D_{t'} \neq D_{t} \,\}. \label{eq:lodo}
\end{align}
LODO therefore forces every fold across a change of imagery, and the two SORT tasks, whose single source already comes from the other dataset, receive identical folds under both protocols. Every choice a method makes on a fold, the ranker variant of \auspex included, is fixed by leaving one source task out inside that fold. With \(\theta^{*}_q\) the configuration committed on sequence \(q\) and \(\theta^{0}_{g}\) the fixed deployment configuration of tracker \(g\), the gain in metric \(m\) is averaged over the sequences of each task and then over tasks, in percentage points,
\begin{equation}
\Delta_{m} = \frac{100}{|\mathcal{T}|}\sum_{t\in\mathcal{T}}\frac{1}{|\mathcal{Q}_t|}\sum_{q\in\mathcal{Q}_t}\bigl[m(q,\theta^{*}_q)-m(q,\theta^{0}_{g_t})\bigr].
\label{eq:gain}
\end{equation}
The best single configuration of a task, \(\theta^{\dagger}_{t}=\arg\max_{\theta}\sum_{q\in\mathcal{Q}_t}\mathrm{IDF1}(q,\theta)\), gains \(\Delta^{\star}=3.10\) points of IDF1, the ceiling for any method that commits one configuration per task. Shifts below half a point on one task are not read as effects, since the load order of numerical libraries alone moves a single task by up to half a point.

\subsubsection{Metrics and Cost}
Accuracy is read through MOTA~\cite{bernardin2008clear} and IDF1~\cite{ristani2016idf1},
\begin{align}
\mathrm{MOTA} &= 1-\frac{\sum_{f}\bigl(\mathrm{FN}_f+\mathrm{FP}_f+\mathrm{IDSW}_f\bigr)}{\sum_{f}\mathrm{GT}_f}, \label{eq:mota}\\
\mathrm{IDF1} &= \frac{2\,\mathrm{IDTP}}{2\,\mathrm{IDTP}+\mathrm{IDFP}+\mathrm{IDFN}}, \label{eq:idf1}
\end{align}
where IDF1, which matches whole trajectories, is primary because association parameters act on identity before detection. Cost is counted as the fraction \(\rho\) of target sequences that carry ground truth, the tracking passes per sequence with a partial run counted by its frames, and the tokens a language model consumes. To read accuracy and labels together, the normalized hypervolume is the rectangle a method spans against the nadir at which every sequence is labelled and nothing is gained,
\begin{equation}
H = (1-\rho)\,\eta, \qquad \eta = \max\bigl(0,\ \Delta_{\mathrm{IDF1}}/\Delta^{\star}\bigr).
\label{eq:hv}
\end{equation}
so that labelling the whole corpus and falling below the fixed configuration both give \(H=0\).

\subsubsection{Agent Missions and Language Models}
The language model above the runtime is exercised on observation missions that model the tasking of a single pass, in which downlink and follow-up observation can serve only a few targets. Given a SatMTB test clip, the agent scans segments, reads the tracks the runtime publishes, and locks and submits three targets, while the runtime commits the configuration on its own. The four missions of Table~\ref{tab:missions} add in turn a sort, a filter, a filter that conflicts with the sort, and a comparison over time.

\begin{table}[t]
\centering
\begin{threeparttable}
\caption{Observation missions posed to the agent.}
\label{tab:missions}
\renewcommand{\arraystretch}{1.15}
\footnotesize
\setlength{\tabcolsep}{3pt}
\begin{tabularx}{\columnwidth}{@{}l >{\raggedright\arraybackslash}X >{\raggedright\arraybackslash}X l@{}}
\toprule
\textbf{Mission} & \textbf{Instruction}\tnote{a} & \textbf{Scenario} & \textbf{Reasoning} \\
\midrule
Area & Lock the three largest targets & Cueing the most significant vessels and aircraft & Sort \\
Transient & Lock the three largest seen in two or more frames & Rejecting likely noise before cueing & Filter \\
Persistence & Lock the three largest seen in three or more frames even over a larger brief one & Holding stable targets for continuous monitoring & Override \\
Growth & Lock the three whose area grows most from early to late frames & Warning of approaching or expanding targets & Compare \\
\bottomrule
\end{tabularx}
\begin{tablenotes}\footnotesize
  \item[a] Each mission grants three detection calls and three tracking slots and ends when the locked targets are submitted.
\end{tablenotes}
\end{threeparttable}
\end{table}

Fifteen open models from six families, from 0.9 to 35 billion parameters within Qwen3.5 alone, are served by Ollama 0.35.0 at temperature zero with a 32K context and at most 1,024 new tokens per turn, each attached by name through the schema of Section~\ref{sec:sys-arch}. Every model reasons in its visible reply before a tool call, with separate thinking modes off or, for GPT-OSS, whose hidden reasoning cannot be disabled, at the lowest effort. Each model runs the four missions on nine sequences drawn evenly from cars, aircraft, and ships. With \(\mathcal{A}\) the submitted set, \(\mathcal{O}\) the targets the agent itself observed, \(R\) the mission applied as a fixed rule, and \(\mathcal{G}\) the three ground-truth targets, execution accuracy, abbreviated EX, and Recall@3, abbreviated R@3, are
\begin{equation}
\mathrm{EX}=\frac{|\mathcal{A}\cap R(\mathcal{O})|}{|R(\mathcal{O})|}, \qquad
\mathrm{R@3}=\frac{|\mathcal{A}\cap\mathcal{G}|}{|\mathcal{G}|}.
\label{eq:ex}
\end{equation}
EX asks whether the model carried out the mission on what it was given, counting ties at the displayed precision as correct, and R@3 scores the area-based missions against ground truth, where applying \(R\) to the reference tracks gives a perception ceiling of 0.67. Every row comes from one server version, since a change of its numerical kernels alone can move the mean EX of a small model by up to 0.47.

\subsubsection{Baselines and Implementation}
The baselines of Table~\ref{tab:overall} span the label-free selectors of Section~\ref{sec:related-labelfree} and the labelled searches of Section~\ref{sec:related-hpo}, with Bayesian optimization on the proxy score of~\cite{davila2026vlmtracker} abbreviated Proxy BO. Methods from other domains receive the inputs of \auspex, detection statistics as meta-features and track statistics as internal measures, and labelled searches stop at twenty evaluations. \auspex uses \(M=3\) and an XGBoost ranker of \(B=300\) trees of depth six at a learning rate of 0.05, and in the comparisons it scores the whole sequence, \(W=n_q\), as the label-free searches do, while the deployed window \(W=25\) is evaluated in Section~\ref{sec:exp-system}. Labelled searches average ten seeds and the label-free searches and LLAMBO three. Trackers run on one CPU core and language models on one NVIDIA RTX 5090 GPU, where the agent layer of Table~\ref{tab:commit} uses Qwen3-VL-4B-Instruct~\cite{bai2025qwen3vl} on textual evidence with greedy decoding.

\subsection{Overall Performance}
\label{sec:exp-main}

Table~\ref{tab:overall} reads every method on accuracy and cost at once. Under LOTO and LODO \auspex gains 2.84 and 2.94 points of IDF1 and 3.21 and 3.29 points of MOTA with no target label, three passes per sequence, and no token, recovers 92 and 95 percent of the ceiling, and attains the highest hypervolume with a MOTA gain no other method exceeds. The baselines fall short in three ways, each rooted in what a method trusts on an unseen target.

Methods that predict without observing are exposed to a change of imagery, and ISAC, MetaOD, and ZAP-HPO all fall below the fixed configuration under LODO because the meta-features of a held-out dataset lie outside the fitted range and nothing observed on the target corrects the extrapolation. Methods that observe through a fixed proxy inherit its bias. Proxy BO, which rewards many long tracks with few gaps, loses MOTA after fifty passes through the false tracks its permissive thresholds create, and an online threshold~\cite{ma2023adaptconf} that never consults source experience loses up to 24.9 points on SatVideoDT, so once a hand-written score departs from identity accuracy, more search drives the tracker further the wrong way. Methods that steer a search with a learned evaluator spend a usable signal inefficiently, since HPOD and HyPer gain markedly less with three passes per sequence than with ten and HPOD falls below the fixed configuration under LODO, where three passes are too few for the search to reach what its evaluator would rank highest and a held-out dataset makes the evaluator extrapolate an absolute score.

Spending target labels moves the bottleneck to the number of evaluations before a search settles. With about sixteen labelled sequences and twenty evaluations, BoundingBox attains the highest IDF1, followed by CTS and ZeroShot ahead of the searches from scratch, so what a method carries from the sources matters more than its rule for the next evaluation, while FSBO and PFNs4BO need about nine passes per target sequence and labels hold the hypervolume of every labelled method to at most 0.86.

Every baseline treats adaptation as a search whose next step depends on a score of the last, so its target cost grows with the precision that score demands. \auspex removes the search from the target at a fixed three passes, and its arbiter makes the relative judgment of Section~\ref{sec:method-arbiter}, which degrades less across datasets than the absolute predictions of MetaOD and HPOD. BoundingBox exceeds it by 0.14 and 0.06 points of IDF1 at the price of about sixteen labelled sequences per task, whereas SQE, the best unlabelled baseline, trails it by 0.68 and 0.78 points.

\begin{figure*}[t]
\centering
\begin{subfigure}[t]{0.329\textwidth}
\centering
\includegraphics[width=\linewidth]{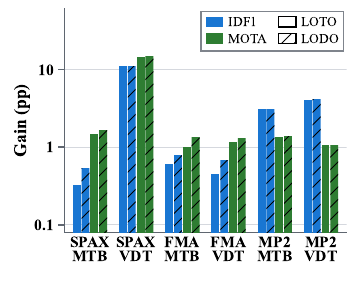}
\caption{Gain per detector and dataset.}
\label{fig:pertask}
\end{subfigure}\hfill
\begin{subfigure}[t]{0.329\textwidth}
\centering
\includegraphics[width=\linewidth]{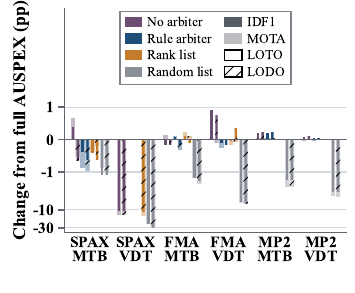}
\caption{Change when one component is altered.}
\label{fig:ablation-structure}
\end{subfigure}\hfill
\begin{subfigure}[t]{0.329\textwidth}
\centering
\includegraphics[width=\linewidth]{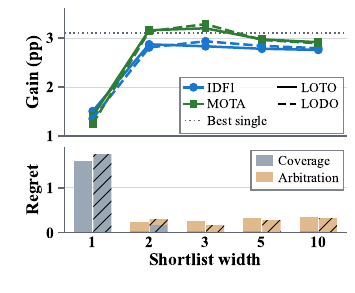}
\caption{Gain against shortlist width.}
\label{fig:ablation-width}
\end{subfigure}
\caption{Sources of the gain of \auspex in IDF1 and MOTA under LOTO and LODO. Blue marks IDF1 and green MOTA, plain bars or solid lines LOTO and hatched bars or dashed lines LODO. (b) Dark narrow bars give IDF1 and light bars MOTA, on an axis linear within half a point of zero and logarithmic beyond. (c) The lower panel stacks the coverage and arbitration regrets below the best single configuration, drawn dotted.}
\label{fig:exp}
\end{figure*}

The gain is uneven across tasks and follows the room that the fixed configuration leaves. In Fig.~\ref{fig:pertask}, on a logarithmic axis with the two SatMTB splits of SPAX combined, SPAX on SatVideoDT, where the best single configuration lies 11.2 points above the fixed one, contributes about eleven points, the two MP2Net tasks with 3.2 and 4.2 points of room contribute 3.1 and 4.1, and the four tasks with at most 1.3 points of room less than one point each. The gain correlates with the room at 0.998 under both protocols and stays positive on every task in both metrics. Without SPAX on SatVideoDT \auspex keeps 1.47 and 1.59 points of IDF1, level with SQE under LOTO and ahead of it under LODO, whereas SQE, HPOD, and HyPer each fall below the fixed configuration on at least one task. The mean is thus dominated by the one task on which a fixed configuration fails badly, the case adaptation exists for, while elsewhere the selector recovers most of the little room left without harming any task.

\begin{figure*}[t]
\centering

  \begin{subfigure}[t]{0.194\textwidth}
  \centering
  \includegraphics[width=\linewidth]{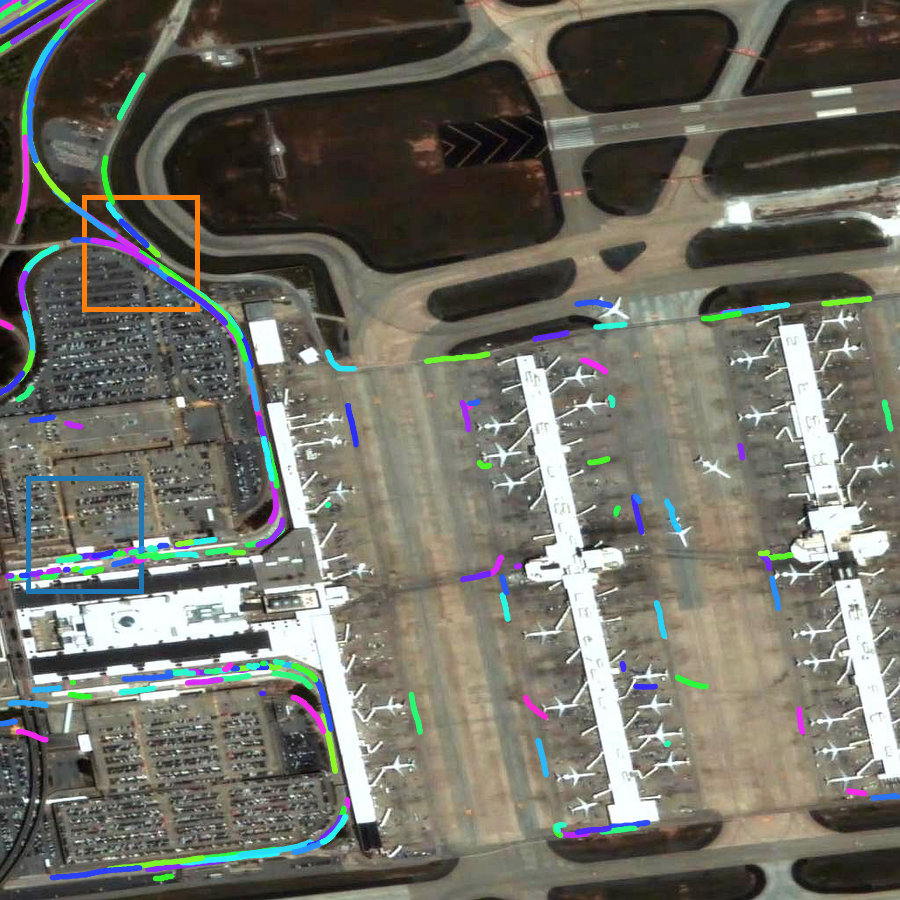}\\[1pt]
  \includegraphics[width=0.492\linewidth]{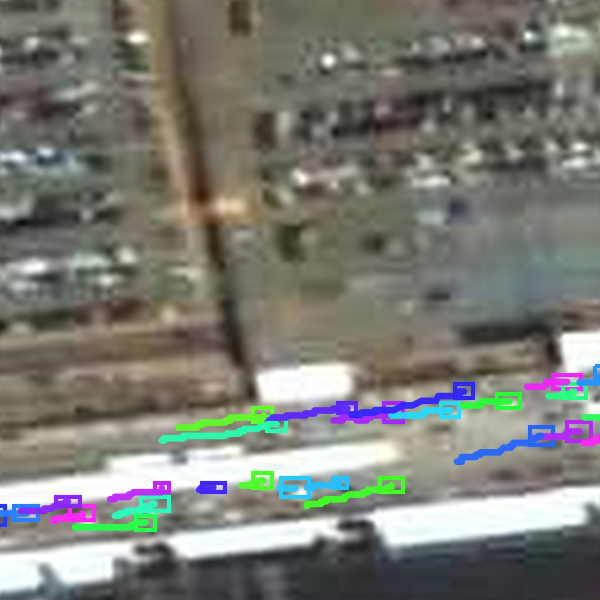}\hfill
  \includegraphics[width=0.492\linewidth]{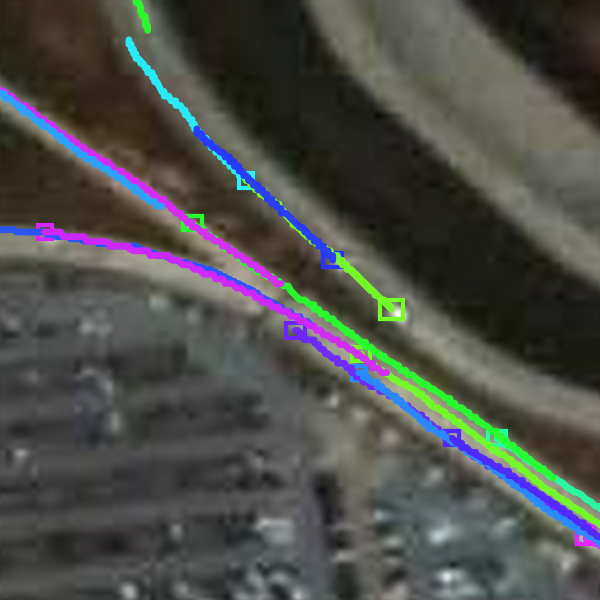}
  \caption{Ground truth}
  \end{subfigure}\hfill
  \begin{subfigure}[t]{0.194\textwidth}
  \centering
  \includegraphics[width=\linewidth]{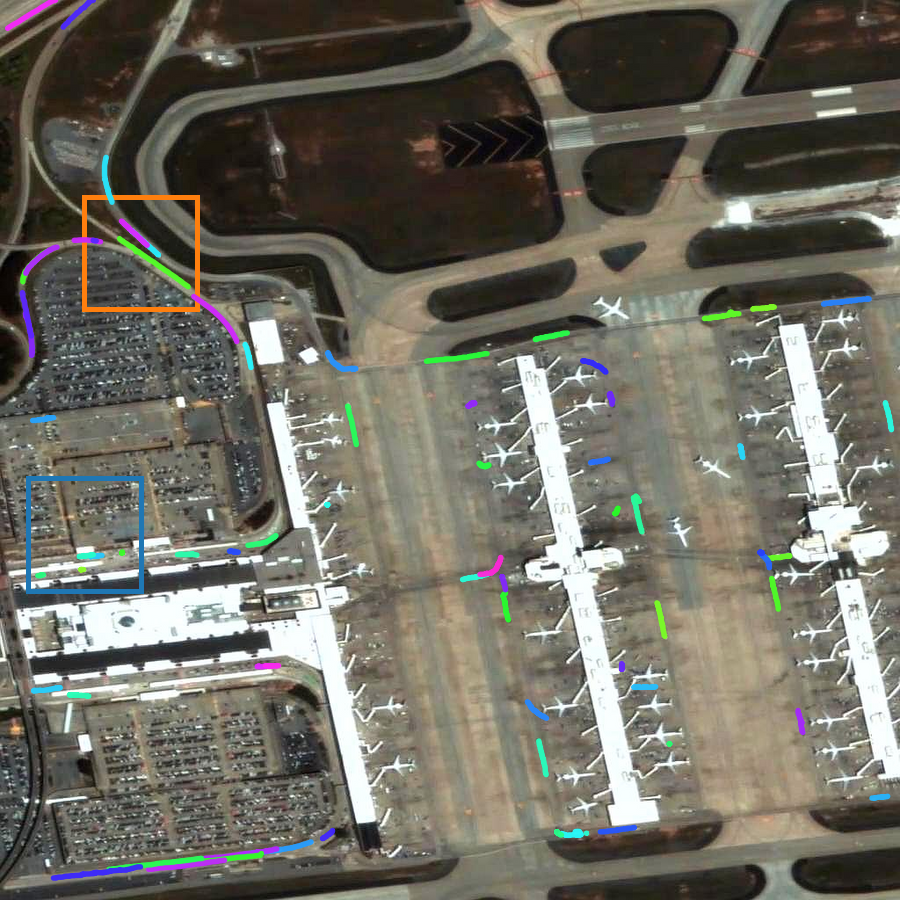}\\[1pt]
  \includegraphics[width=0.492\linewidth]{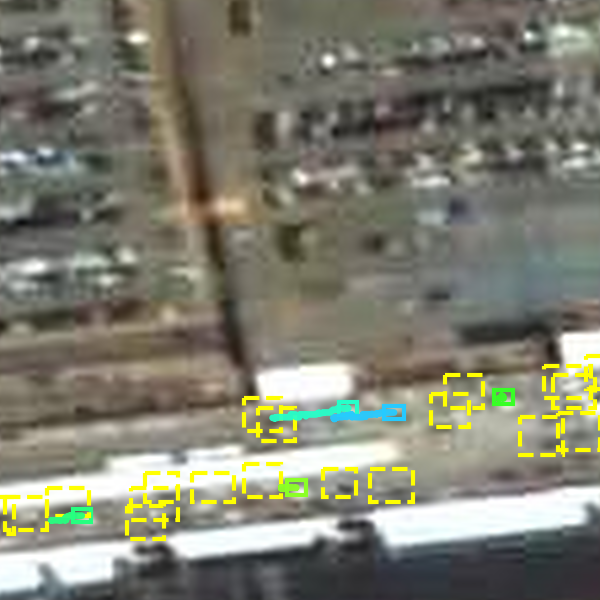}\hfill
  \includegraphics[width=0.492\linewidth]{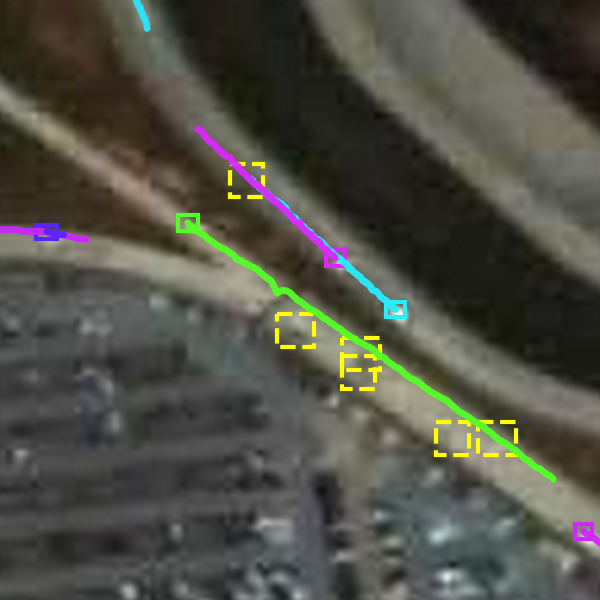}
  \caption{Fixed configuration}
  \end{subfigure}\hfill
  \begin{subfigure}[t]{0.194\textwidth}
  \centering
  \includegraphics[width=\linewidth]{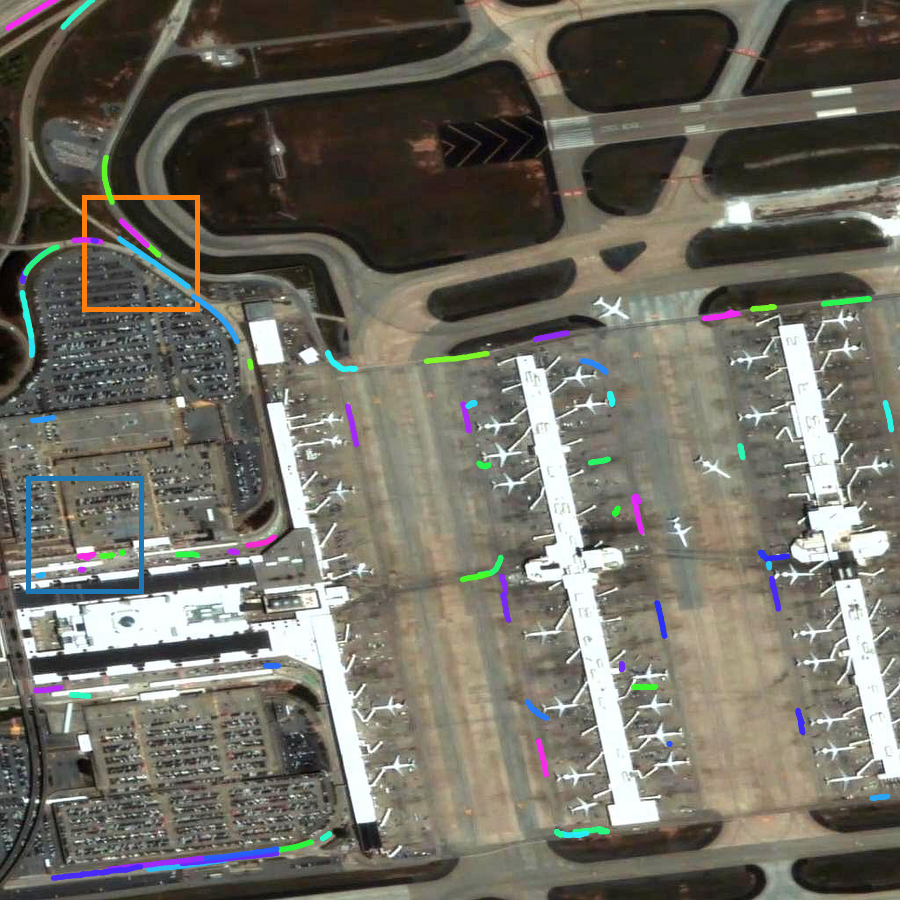}\\[1pt]
  \includegraphics[width=0.492\linewidth]{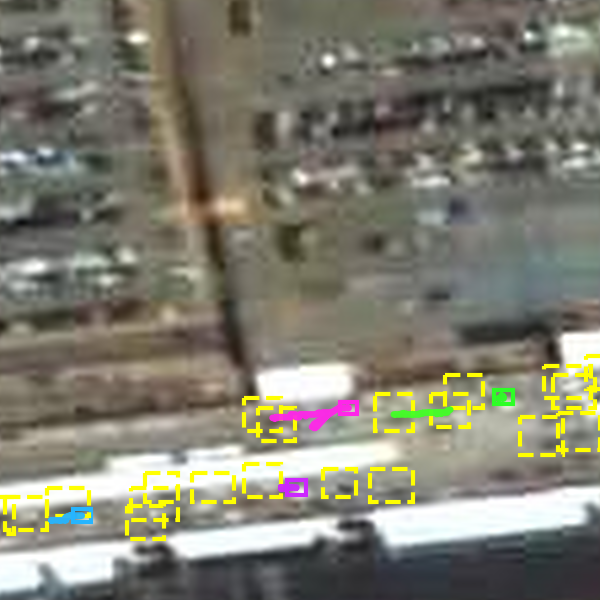}\hfill
  \includegraphics[width=0.492\linewidth]{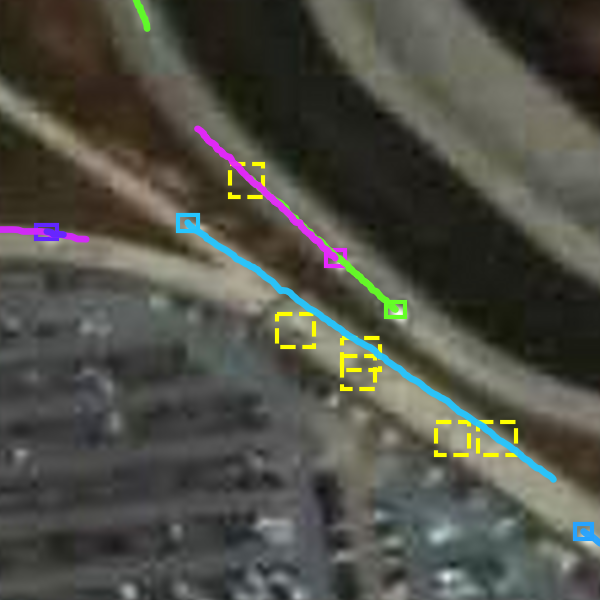}
  \caption{SQE~\cite{huang2020sqe}}
  \end{subfigure}\hfill
  \begin{subfigure}[t]{0.194\textwidth}
  \centering
  \includegraphics[width=\linewidth]{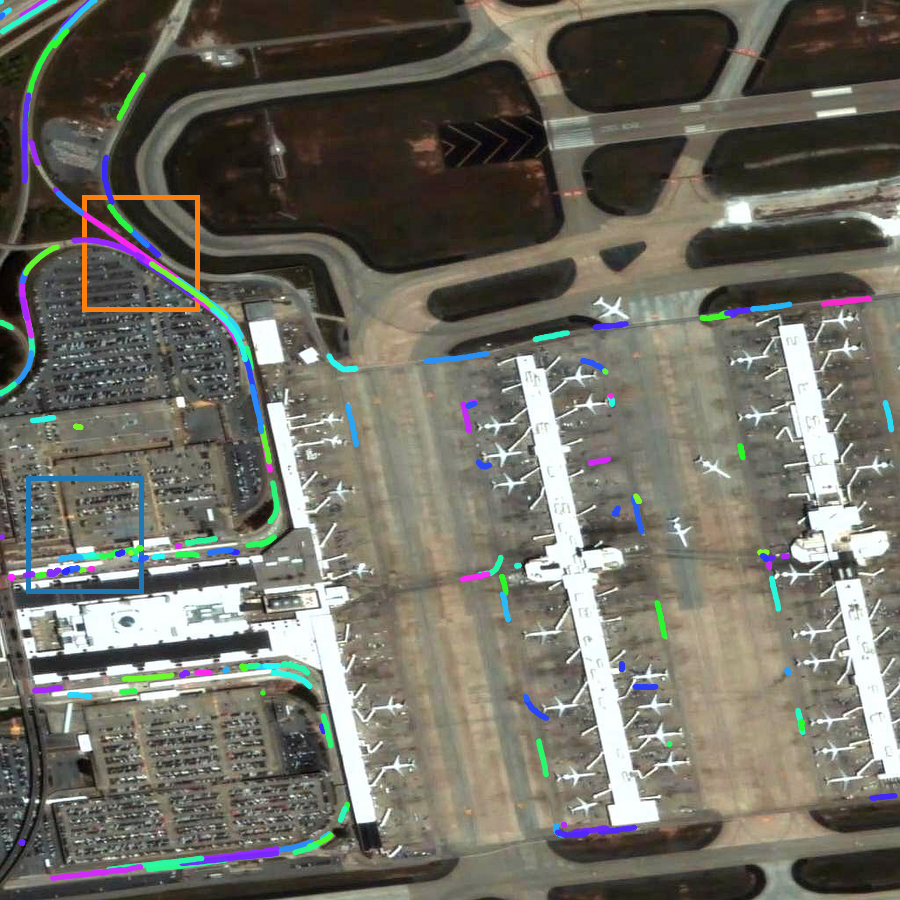}\\[1pt]
  \includegraphics[width=0.492\linewidth]{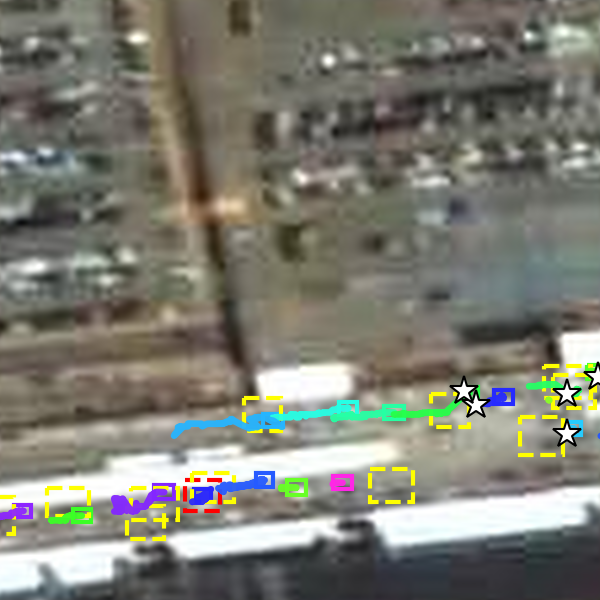}\hfill
  \includegraphics[width=0.492\linewidth]{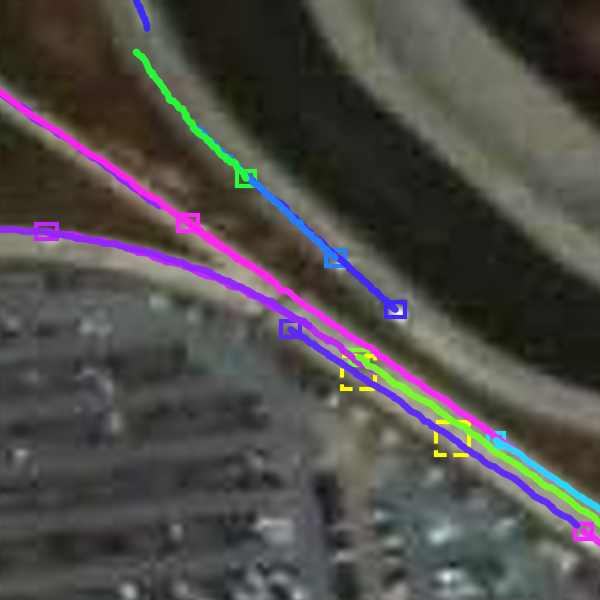}
  \caption{\auspex}
  \end{subfigure}\hfill
  \begin{subfigure}[t]{0.194\textwidth}
  \centering
  \includegraphics[width=\linewidth]{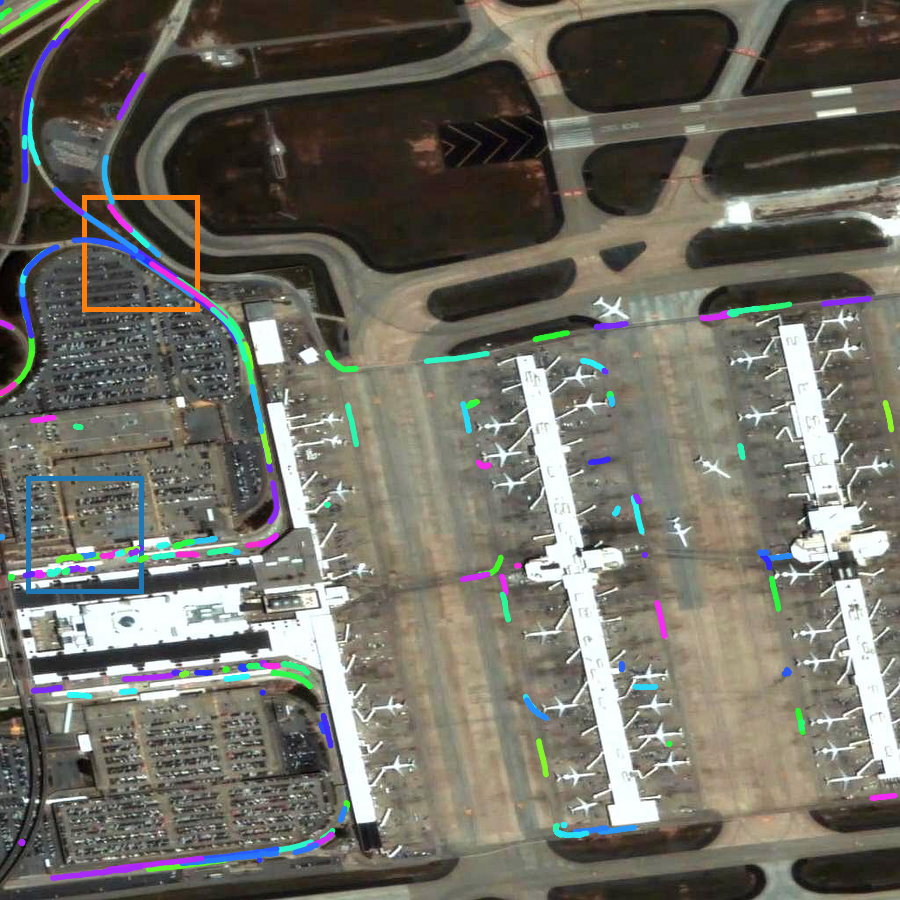}\\[1pt]
  \includegraphics[width=0.492\linewidth]{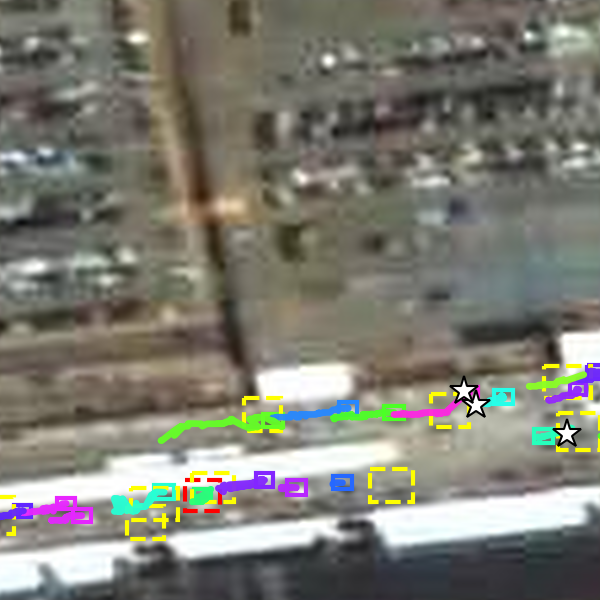}\hfill
  \includegraphics[width=0.492\linewidth]{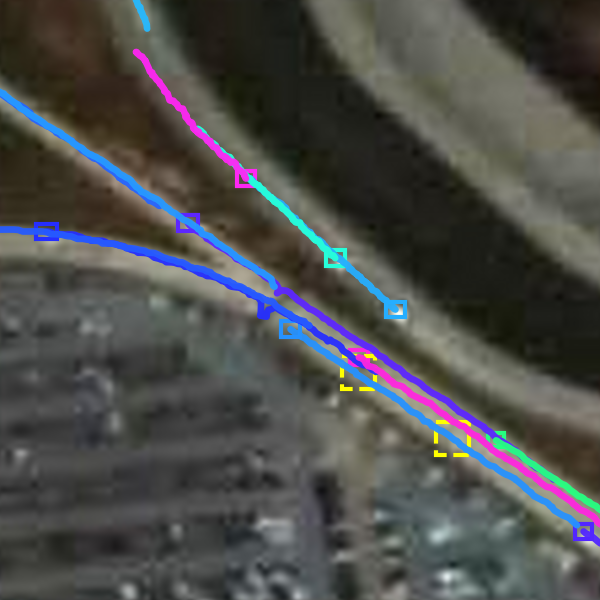}
  \caption{Best configuration}
  \end{subfigure}
\caption{Trajectories on video 08 of SatVideoDT with SPAX detections over frames 294 to 443. Each column shows the full frame above the two regions outlined in blue and orange. Lines trace the last 150 frames and solid boxes mark the tracks in frame 443, both colored by identity, while yellow and red dashed boxes mark missed targets and false tracks and white stars mark identity switches. The regions are those in which the baselines miss the most targets that \auspex tracks, and the best configuration is chosen for this sequence with ground truth.}
\label{fig:qual}
\end{figure*}

Fig.~\ref{fig:qual} follows one sequence of that task, video 08 of SatVideoDT under SPAX, whose gain of 18.40 points ranks seventeenth of its seventy sequences. Every column tracks the same detections, so each difference arises from the configuration alone. The fixed configuration starts a track only above a score of 0.30 and leaves most vehicles unconfirmed at an IDF1 of 41.67 percent, and SQE keeps the same threshold at 39.63 percent, since a score on the geometry of existing tracks is not penalized by a target never tracked. \auspex lowers the threshold to 0.25, as the best configuration for the sequence does, and reaches 60.07 percent against its 62.33. The recovered coverage brings more false tracks and identity switches, yet the missed targets fall by about a third and outweigh both, so the gain comes from undoing a deployment setting that suppresses the very targets a mission must follow.

\begin{table*}[t]
\centering
\begin{threeparttable}
\caption{Average gain over the fixed configuration against state-of-the-art baselines.}
\label{tab:overall}
\renewcommand{\arraystretch}{1.15}
\footnotesize
\setlength{\tabcolsep}{0pt}
\begin{tabular*}{\textwidth}{@{\extracolsep{\fill}}l c r r r cc cc cc@{}}
\toprule
\multirow{2}{*}{\textbf{Method}\tnote{a}}
  & \multirow{2}{*}{\textbf{Year}}
  & \multicolumn{3}{c}{\textbf{Cost}\tnote{b}}
  & \multicolumn{2}{c}{\textbf{IDF1 gain}}
  & \multicolumn{2}{c}{\textbf{MOTA gain}}
  & \multicolumn{2}{c}{\textbf{Normalized hypervolume \(H\)}} \\
\cmidrule(lr){3-5}\cmidrule(lr){6-7}\cmidrule(lr){8-9}\cmidrule(lr){10-11}
  & & Labels & Passes & Tokens
  & LOTO & LODO & LOTO & LODO & LOTO & LODO \\
\midrule
\rowcolor{gray!10}
\multicolumn{11}{c}{\textbf{No target labels}} \\
ISAC$^{*}$~\cite{kadioglu2010isac}
  & 2010 & 0 & 0 & 0
  & $\phantom{-}$1.04 & $-$0.71 & $\phantom{-}$0.94 & $-$0.84 & 0.34 & 0 \\
MetaOD$^{*}$~\cite{zhao2021metaod}
  & 2021 & 0 & 0 & 0
  & $-$2.87 & $-$6.27 & $-$3.19 & $-$6.34 & 0 & 0 \\
ZAP-HPO$^{*}$~\cite{ozturk2022zap}
  & 2022 & 0 & 0 & 0
  & $-$0.50 & $-$3.08 & $-$1.60 & $-$4.00 & 0 & 0 \\
\addlinespace[3pt]
SQE$^{*}$~\cite{huang2020sqe}
  & 2020 & 0 & 3 & 0
  & $\phantom{-}$2.16 & $\phantom{-}$2.16 & $\phantom{-}$2.14 & $\phantom{-}$2.13 & 0.70 & 0.70 \\
Proxy BO$^{*}$~\cite{davila2026vlmtracker}
  & 2026 & 0 & 50 & 0
  & $\phantom{-}$0.60 & $\phantom{-}$0.60 & $-$2.05 & $-$2.05 & 0.19 & 0.19 \\
\addlinespace[3pt]
HPOD$^{*}$~\cite{zhao2024hpod}
  & 2024 & 0 & 3 & 0
  & $\phantom{-}$1.15 & $-$2.48 & $\phantom{-}$0.94 & $-$2.82 & 0.37 & 0 \\
HyPer$^{*}$~\cite{ding2024hyper}
  & 2024 & 0 & 3 & 0
  & $\phantom{-}$1.37 & $\phantom{-}$1.36 & $\phantom{-}$1.55 & $\phantom{-}$1.37 & 0.44 & 0.44 \\
\rowcolor{gray!10}
\multicolumn{11}{c}{\textbf{Target labels required}} \\
Random search~\cite{bergstra2012random}
  & 2012 & 16.4/15.7 & 3.6/3.4 & 0
  & $\phantom{-}$2.35 & $\phantom{-}$2.33 & $\phantom{-}$2.61 & $\phantom{-}$2.65 & 0.62 & 0.62 \\
GP-BO~\cite{snoek2012practical}
  & 2012 & 16.4/15.7 & 3.6/3.4 & 0
  & $\phantom{-}$2.85 & $\phantom{-}$2.83 & $\phantom{-}$2.97 & $\phantom{-}$2.91 & 0.76 & 0.76 \\
Successive halving~\cite{jamieson2016non}
  & 2016 & 16.4/15.7 & 3.6/3.4 & 0
  & $\phantom{-}$2.78 & $\phantom{-}$2.78 & $\phantom{-}$2.91 & $\phantom{-}$2.91 & 0.74 & 0.74 \\
\addlinespace[3pt]
ZeroShot~\cite{wistuba2015smfo}
  & 2015 & 5.1 & 1.1 & 0
  & $\phantom{-}$2.76 & $\phantom{-}$2.80 & $\phantom{-}$2.90 & $\phantom{-}$2.89 & 0.84 & 0.86 \\
ZeroShot~\cite{wistuba2015smfo}
  & 2015 & 9.6 & 2.0 & 0
  & $\phantom{-}$2.87 & $\phantom{-}$2.87 & $\phantom{-}$2.98 & $\phantom{-}$3.00 & 0.83 & 0.83 \\
ZeroShot~\cite{wistuba2015smfo}
  & 2015 & 16.4/15.7 & 3.6/3.4 & 0
  & $\phantom{-}$2.94 & $\phantom{-}$2.95 & $\phantom{-}$3.15 & $\phantom{-}$3.17 & 0.78 & 0.79 \\
BoundingBox~\cite{perrone2019boundingbox}
  & 2019 & 16.4/15.7 & 3.6/3.4 & 0
  & $\phantom{-}$\textbf{2.98} & $\phantom{-}$\textbf{3.00} & $\phantom{-}$\textbf{3.21} & $\phantom{-}$3.23 & 0.79 & 0.80 \\
CTS~\cite{salinas2020cts}
  & 2020 & 16.4/15.7 & 3.6/3.4 & 0
  & $\phantom{-}$2.96 & $\phantom{-}$2.96 & $\phantom{-}$3.16 & $\phantom{-}$3.09 & 0.78 & 0.79 \\
FSBO~\cite{wistuba2021fsbo}
  & 2021 & 16.4/15.7 & 3.6/3.4 & 0
  & $\phantom{-}$2.64 & $\phantom{-}$2.62 & $\phantom{-}$2.87 & $\phantom{-}$2.86 & 0.70 & 0.70 \\
\addlinespace[3pt]
PFNs4BO~\cite{muller2023pfns4bo}
  & 2023 & 16.4/15.7 & 3.6/3.4 & 0
  & $\phantom{-}$2.89 & $\phantom{-}$2.87 & $\phantom{-}$2.93 & $\phantom{-}$2.92 & 0.76 & 0.77 \\
LLAMBO$^{*}$~\cite{liu2024llambo}
  & 2024 & 58 & 11.8 & \(5.0\times 10^{5}\)
  & $\phantom{-}$2.78 & $\phantom{-}$2.78 & $\phantom{-}$3.14 & $\phantom{-}$3.14 & 0.37 & 0.37 \\
\midrule
\textbf{\auspex (Ours)}
  & 2026 & 0 & 3 & 0
  & $\phantom{-}$2.84 & $\phantom{-}$2.94 & $\phantom{-}$\textbf{3.21} & $\phantom{-}$\textbf{3.29}
  & \textbf{0.91} & \textbf{0.95} \\
\textcolor{gray}{Best target configuration}\tnote{c}
  & \textcolor{gray}{\textemdash} & \textcolor{gray}{97} & \textcolor{gray}{199} & \textcolor{gray}{0}
  & \textcolor{gray}{$\phantom{-}$3.10} & \textcolor{gray}{$\phantom{-}$3.10}
  & \textcolor{gray}{$\phantom{-}$3.38} & \textcolor{gray}{$\phantom{-}$3.38}
  & \textcolor{gray}{0} & \textcolor{gray}{0} \\
\bottomrule
\end{tabular*}
\begin{tablenotes}\footnotesize
  \item[a] Methods marked $^{*}$ are reimplemented and transferred to this domain and task.
  \item[b] Labelled target sequences per task, tracking passes per sequence, and tokens per task; a pair gives LOTO, then LODO.
  \item[c] One configuration per task, chosen by running every configuration on every labelled target sequence; an upper bound in gray, excluded from the bold best, with \(H=0\) because \(\rho=1\).
\end{tablenotes}
\end{threeparttable}
\end{table*}

\subsection{Algorithm-Level Ablations}
\label{sec:exp-ablation}

\subsubsection{Structure of the Selector}
Fig.~\ref{fig:ablation-structure} alters one component at a time against full \auspex. Replacing the learned ranker with the rule that keeps the configuration emitting the most boxes moves the mean by at most 0.13 points of IDF1. Withdrawing arbitration, so that every sequence receives the head of the list, costs 1.33 and 1.58 points, entirely through SPAX on SatVideoDT. Ordering the shortlist by mean source rank instead of coverage costs 1.67 points under LODO but nothing under LOTO, where the sources resemble the held-out task, and a random list lowers all seven tasks. What the shortlist contains thus decides the gain, any reasonable arbiter suffices once the list covers the region the target requires, and the learned ranker is retained as marginally better than the rule at no extra cost.

\subsubsection{Width of the Shortlist}
Fig.~\ref{fig:ablation-width} varies the width \(M\) and splits the IDF1 gap to the best single configuration into the two regrets of \eqref{eq:regret}. Widening from one configuration to two cuts the coverage regret from 1.59 and 1.75 points to 0.03 and 0.17, at three it is 0.03 under both protocols, and beyond three only the arbitration regret grows as more candidates compete. \(M=3\) is therefore the narrowest width that covers the target under both protocols, and widening further only hands the arbiter more chances to err.

\subsection{System-Level Ablations}
\label{sec:exp-system}

The ablations below hold the algorithm fixed and alter the runtime \corvus of Section~\ref{sec:system}, asking when it commits, which layer commits, and whether the language model above can be exchanged, questions that arise only once selection runs inside an agent on a live stream and so test the framework rather than the selector.

\subsubsection{Granularity of Commitment}
The upper block of Table~\ref{tab:commit} asks how finely and how early to commit. One configuration per task, chosen without labels from the same three passes, already recovers most of the gain, but the oracle over the shortlist rises when it may choose per sequence, so the remaining headroom lies at the finer granularity. Committing from the first 25 frames matches the full-sequence IDF1 under LOTO and trails it by 0.28 points under LODO at 1.25 passes, as \eqref{eq:cost} predicts, whereas emitting the leading configuration during those frames and switching afterward costs more. Recommitting every 50 frames loses about a point, and even an oracle free to switch every 50 frames reaches only about 2.70, because every switch severs identities. The commitment is therefore a sequence-level decision taken once from a short prefix, the event on which the runtime fires the skill, and it releases tracks after 2.5 s at a cost of at most 0.28 points.

\begin{table*}[t]
\centering
\begin{threeparttable}
\caption{Ablation of commitment granularity and commitment layer.}
\label{tab:commit}
\renewcommand{\arraystretch}{1.15}
\footnotesize
\setlength{\tabcolsep}{0pt}
\begin{tabular*}{\textwidth}{@{\extracolsep{\fill}}l r r c cc cc cc@{}}
\toprule
\multirow{2}{*}{\textbf{Variant}\tnote{a}}
  & \multicolumn{3}{c}{\textbf{Cost per sequence}\tnote{b}}
  & \multicolumn{2}{c}{\textbf{IDF1 gain}}
  & \multicolumn{2}{c}{\textbf{MOTA gain}}
  & \multicolumn{2}{c}{\textbf{Oracle IDF1 gain}\tnote{c}} \\
\cmidrule(lr){2-4}\cmidrule(lr){5-6}\cmidrule(lr){7-8}\cmidrule(lr){9-10}
  & Passes & Tokens & Switches
  & LOTO & LODO & LOTO & LODO & LOTO & LODO \\
\midrule
\textbf{\auspex}
  & 3 & 0 & 0
  & \textbf{2.84} & \textbf{2.94} & 3.21 & 3.29 & 3.34 & 3.35 \\
\rowcolor{gray!10}
\multicolumn{10}{c}{\textbf{Commitment granularity}} \\
Task
  & 3 & 0 & 0
  & 2.78 & 2.88 & 3.14 & \textbf{3.33} & 3.08 & 3.08 \\
Prefix, deferred
  & 1.25 & 0 & 0
  & 2.82 & 2.66 & \textbf{3.25} & 3.03 & 3.34 & 3.35 \\
Prefix, immediate
  & 1.25 & 0 & 0.87/0.79
  & 2.57 & 2.46 & 2.98 & 2.81 & 3.34 & 3.35 \\
Window
  & 3 & 0 & 1.82/1.69
  & 1.90 & 1.90 & 2.71 & 2.70 & 2.70 & 2.71 \\
\rowcolor{gray!10}
\multicolumn{10}{c}{\textbf{Commitment layer}} \\
Perception, fixed
  & 1 & 0 & 0
  & 0.00 & 0.00 & 0.00 & 0.00 & \textemdash & \textemdash \\
Skill, rule
  & 3 & 0 & 0
  & 2.77 & 2.81 & 2.98 & 3.04
  & \multirow{4}{*}{3.34} & \multirow{4}{*}{3.35} \\
Agent, zero-shot
  & 3 & 1\,618 & 0
  & 2.42 & 2.55 & 2.58 & 2.62 & & \\
Agent, voting
  & 3 & 4\,854 & 0
  & 2.50 & 2.61 & 2.61 & 2.68 & & \\
Agent, few-shot
  & 3 & 13\,038 & 0
  & 2.82 & 2.33 & 2.97 & 2.40 & & \\
\bottomrule
\end{tabular*}
\begin{tablenotes}\footnotesize
  \item[a] \auspex scores the whole sequence and commits one configuration per sequence at the skill layer, and each block alters only one of these two factors. Prefix variants decide on the first 25 frames and Window every 50 frames; Agent variants hand the evidence of the ranker to Qwen3-VL-4B-Instruct.
  \item[b] Tracking passes including the one whose output is kept, prompt tokens, and configuration changes inside a sequence; a pair gives LOTO, then LODO.
  \item[c] Best gain at the same granularity when the choice among the same three shortlisted configurations is made with ground truth.
\end{tablenotes}
\end{threeparttable}
\end{table*}

\subsubsection{Layer of Commitment}
The lower block keeps the three candidates and the evidence of the ranker fixed and moves only the layer that commits. At the perception layer the gain is zero by construction, and at the skill layer even a hand-written rule in place of the ranker nearly matches it. At the agent layer, where a language model reads the same evidence and names the configuration, a single call trails the skill at about 1,600 tokens per sequence and voting over three rotations of the candidate order narrows the gap only slightly at three times the tokens, while three source examples lift it to 2.82 points under LOTO but drop it to 2.33 under LODO, where the examples teach a preference the target does not share. Handing the model only the sequences of small ranker margin moves the mean by at most 0.04 points. Moving the commitment from perception to the skill thus produces the adaptation, whereas moving it onward to the agent adds tokens and a dependence on source examples without adding accuracy, the placement that \eqref{eq:placement} prescribes.

\subsubsection{Interchangeability of the Language Model}
\label{sec:exp-models}
Table~\ref{tab:models} exchanges the language model while the skill commits. In every run that reached the commitment the configuration committed on each sequence is identical across all fifteen models, so neither the perception every model reads nor its ceiling of 0.67 depends on the model. EX rises with scale, within Qwen3.5 from 0.14 at 0.9 billion parameters to 0.99 at 27 billion, with models of nine billion or more from the other families in the same band and the growth mission separating models most sharply, even Qwen3.8 at 27 billion trailing its predecessor on it, while two mixture-of-experts models with 3 and 3.8 billion active parameters reach the same band, so capability follows the stored size and latency the active size. R@3 stops at the perception ceiling, which places the end-to-end bottleneck in perception once a model of about nine billion parameters is attached, a trend of this suite and prompt rather than a law of scale. Six families differing in architecture, native tool format, and nearly fortyfold in scale thus attach through one schema without changing prompts, tools, or runtime, so a weak model fails the mission without degrading the tracking and a newer model or skill can be uplinked on its own schedule.

\begin{figure}[t]
\centering

\begin{subfigure}[t]{0.326\columnwidth}
\centering
  \includegraphics[width=\linewidth]{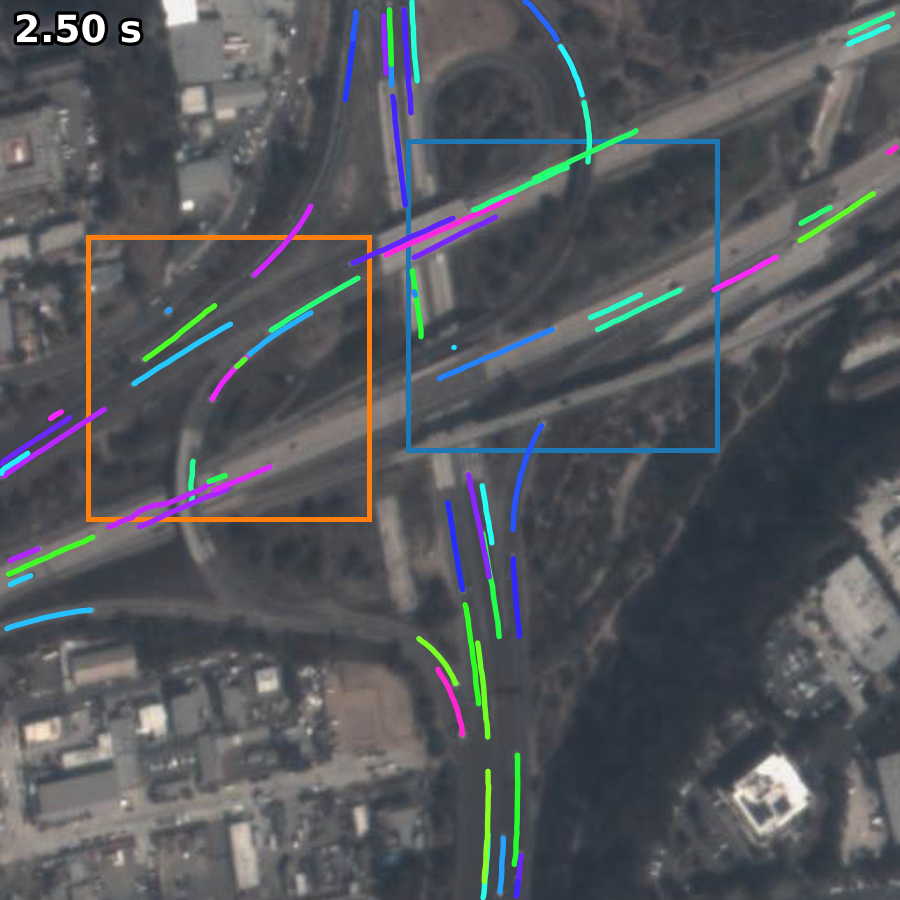}\\[1pt]
  \includegraphics[width=0.49\linewidth]{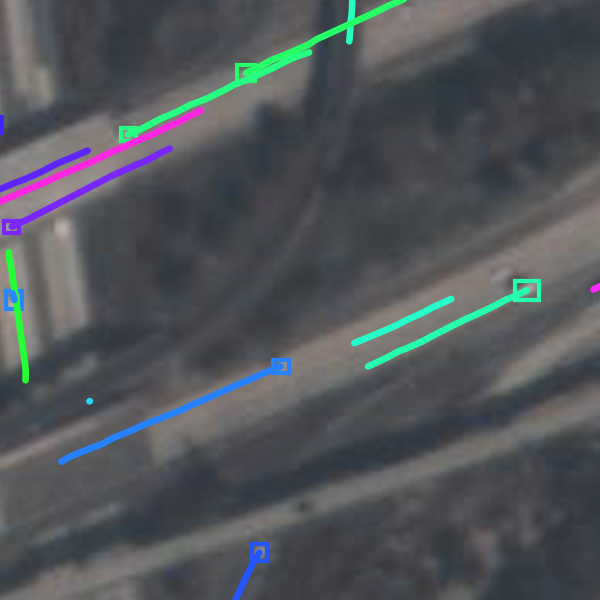}\hfill
  \includegraphics[width=0.49\linewidth]{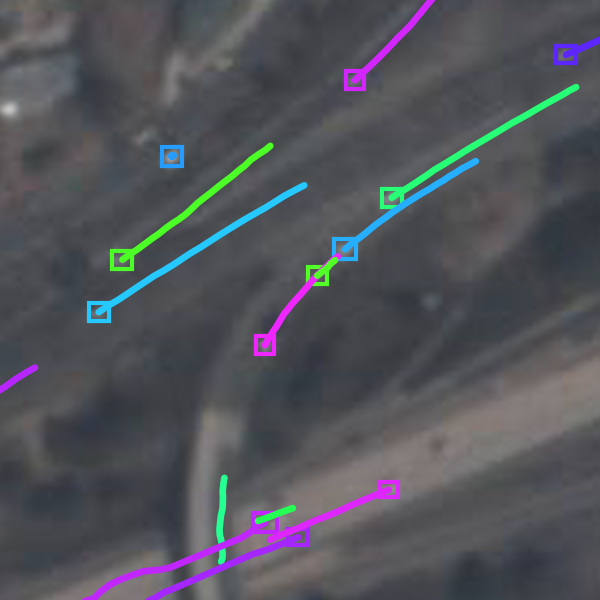}
\caption{Commitment}
\label{fig:tl-commit}
\end{subfigure}\hfill
\begin{subfigure}[t]{0.326\columnwidth}
\centering
  \includegraphics[width=\linewidth]{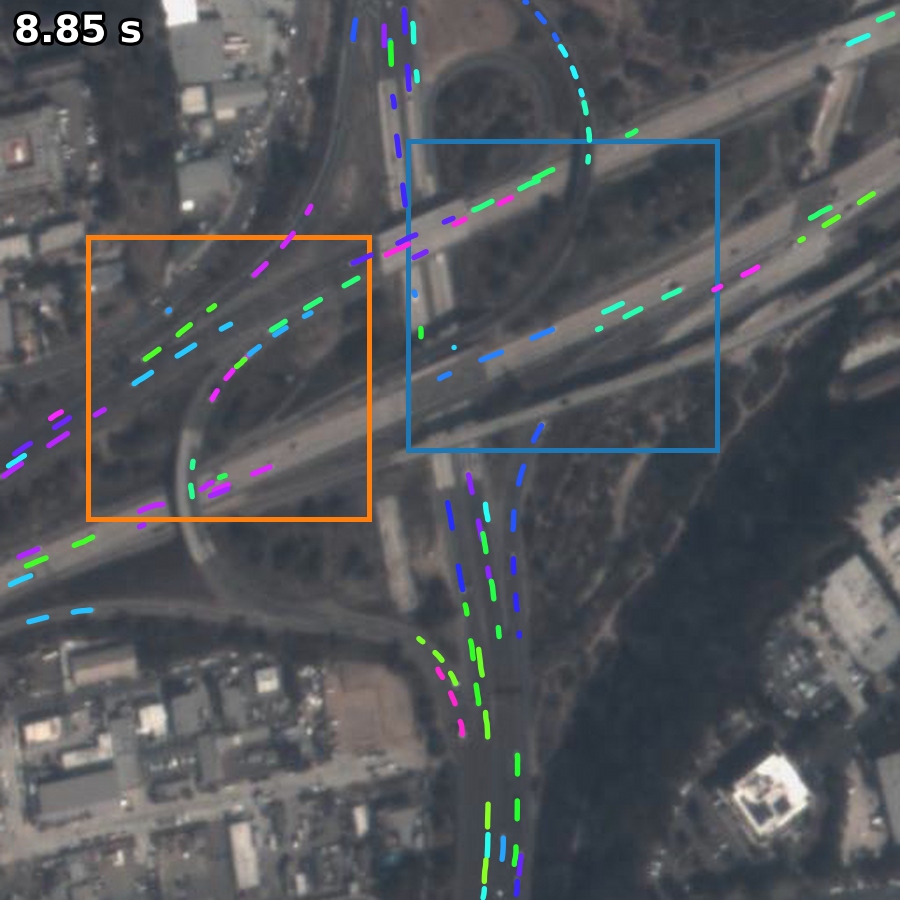}\\[1pt]
  \includegraphics[width=0.49\linewidth]{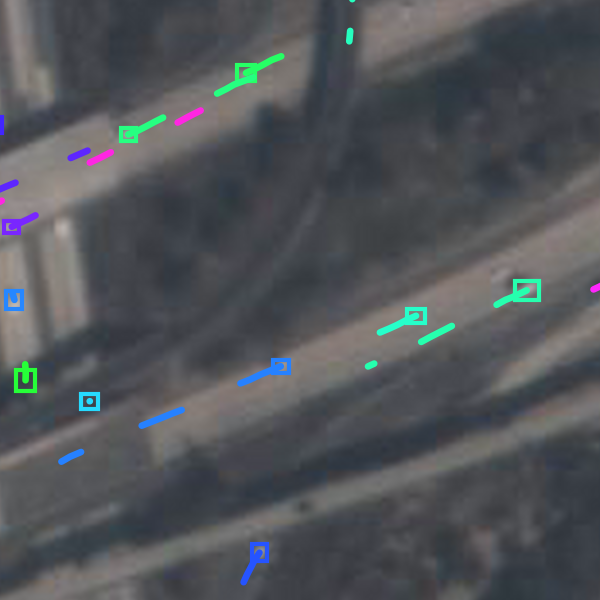}\hfill
  \includegraphics[width=0.49\linewidth]{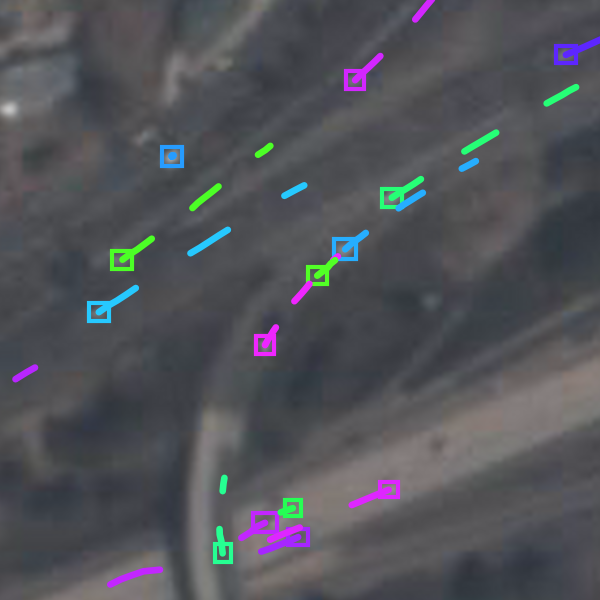}
\caption{Scanned evidence}
\label{fig:tl-scan}
\end{subfigure}\hfill
\begin{subfigure}[t]{0.326\columnwidth}
\centering
  \includegraphics[width=\linewidth]{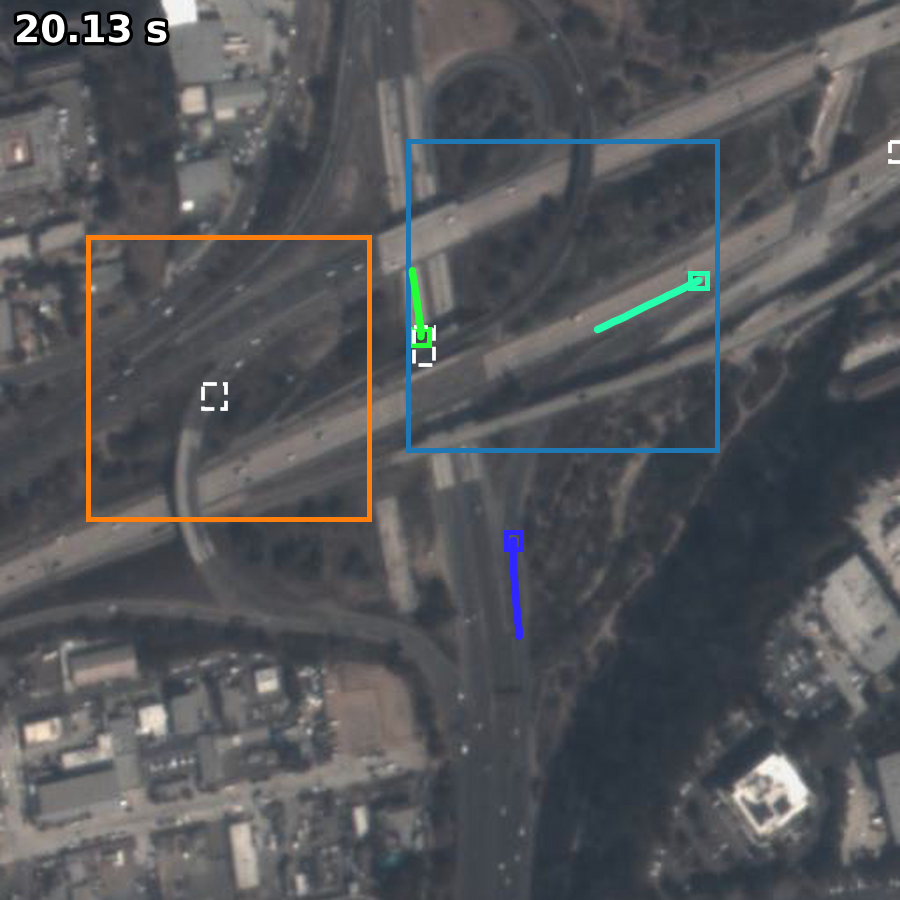}\\[1pt]
  \includegraphics[width=0.49\linewidth]{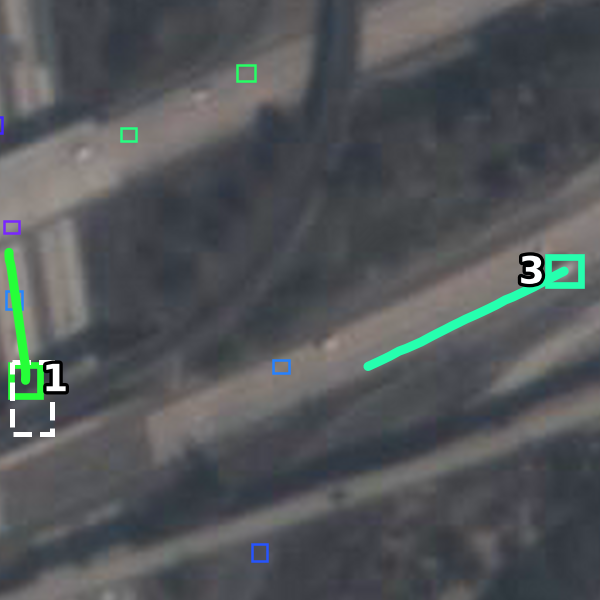}\hfill
  \includegraphics[width=0.49\linewidth]{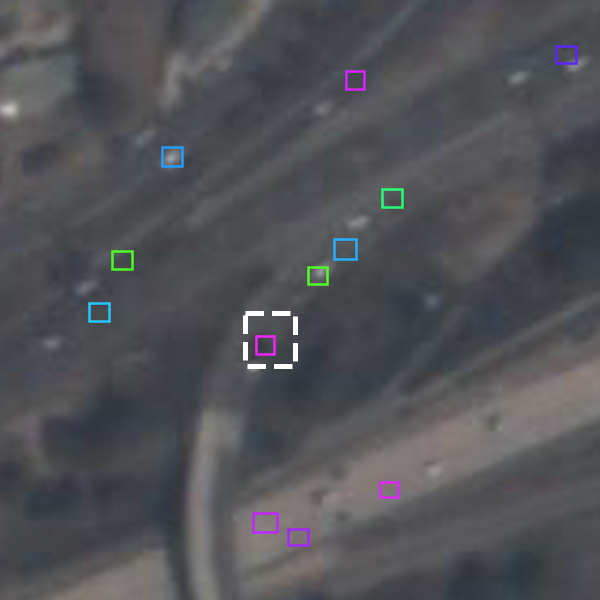}
\caption{Submission}
\label{fig:tl-submit}
\end{subfigure}
\caption{One observation of the persistence mission by Qwen3.5-9B on a SatMTB sequence, timed on the 10 fps frame clock. Each column shows the full frame above the regions outlined in blue and orange, at the commitment from the 25-frame prefix, at the return of the third scan with only the three scanned windows drawn, and at the submission with the locked targets numbered in order of selection and the three largest targets by ground truth in white dashed boxes. Lines and boxes are colored by identity.}
\label{fig:timeline}
\end{figure}

The models also bound the case for an agent layer. Offered the configuration call as a tool on the persistence mission, Qwen3.5 models of 2 to 27 billion parameters invoke it on at most three of nine sequences with no trend in scale, and their R@3 falls 0.19 to 0.44 below the skill layer at every scale. Over 36 paired runs the skill layer wins 21, ties 13, and loses 2, \(p\approx 7\times 10^{-5}\) by a two-sided sign test, and the 2 and 4 billion parameter models committing through the skill, at 0.52 and 0.59, both exceed the 27 billion parameter model left to call the tool, at 0.37. Although the tool arm also lacks the output gate, its dominant failure is a model that never asks, which no tested scale repairs, so scale improves how a mission is executed but does not substitute for committing below the agent.
Fig.~\ref{fig:timeline} follows one run of the persistence mission by the nine-billion-parameter Qwen3.5 on a SatMTB sequence. The first scan, requested at 1.42 s, waits at the output gate until the skill commits at 2.50 s from the prefix of Fig.~\ref{fig:tl-commit}, with a margin of 0.13 below the threshold of 0.5 recorded as a low-margin event in the downlink. The three scans cover five frames each, so the evidence of Fig.~\ref{fig:tl-scan} is fragmentary, and the agent locks the three targets of largest observed area between 11.72 and 17.24 s and submits at 20.13 s, an EX of 1.00. Its recall of 0.33 equals the perception ceiling of the sequence, because the detector draws the first locked target at about 2.7 times its true area and a ground-truth target in the orange region at less than a third of its own, so the agent executed the mission exactly on what perception reported and the remaining error lies in the boxes it was given.

\subsection{Onboard Deployment}
\label{sec:exp-edge}

\begin{figure}[t]
\centering
\includegraphics[width=\columnwidth]{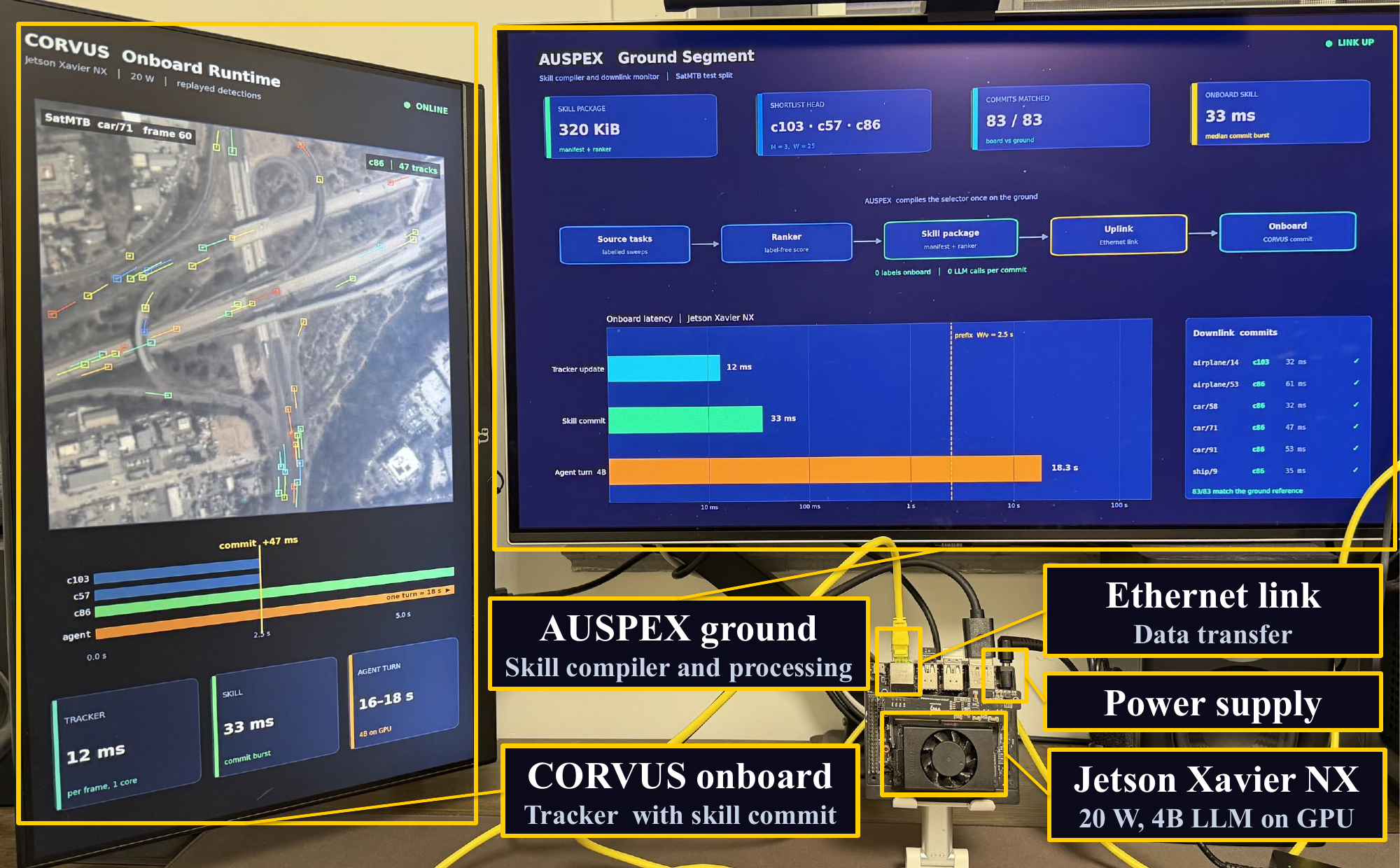}
\caption{Onboard deployment of \corvus on an NVIDIA Jetson Xavier NX, with the ground segment of \auspex on the host and the Ethernet link standing for the uplink and downlink.}
\label{fig:edge}
\end{figure}

Measured first on the workstation of Section~\ref{sec:exp-setup}, the terms of \eqref{eq:load} show the two extra trackers of the prefix adding at most about 13 ms to the frame interval of 100 ms for 25 frames and the skill burst staying below about 22 ms, whereas one turn of a nine-billion-parameter model takes 2.8 to 3.5 s on the GPU, longer than the whole prefix, so \(\lambda_{\mathrm{A}}\ge W/\nu\) holds and the agent cannot commit in time.
The same holds on an NVIDIA Jetson Xavier NX in its 20 W mode, shown in Fig.~\ref{fig:edge}, where \corvus replays the detector output of the 83 SatMTB test sequences with the package compiled on the ground and commits the same configuration as the workstation on every sequence.
One tracker update takes 12 ms per frame at the median on one CPU core, the three trackers of the prefix process its 25 frames in at most 2.07 s on three cores, within the 2.5 s the frames take to arrive, and the skill burst takes 33 ms at the median and 61 ms at most.
One turn of a four-billion-parameter model on the board GPU takes 15.6 to 18.4 s, more than six times the prefix.
The commitment therefore keeps pace with the frame clock on an edge module, while the agent stays off the critical path by a wider margin than on the workstation.

\begin{table}[t]
\centering
\begin{threeparttable}
\caption{Ablation of language-model family and scale.}
\label{tab:models}
\renewcommand{\arraystretch}{1.1}
\footnotesize
\setlength{\tabcolsep}{0pt}
\begin{tabular*}{\columnwidth}{@{\extracolsep{\fill}}l cc ccc cc@{}}
\toprule
\multirow{2}{*}{\textbf{Model}}
  & \multicolumn{2}{c}{\textbf{Params (B)}\tnote{a}}
  & \multicolumn{3}{c}{\textbf{EX}\tnote{b}}
  & \multirow{2}{*}{\textbf{R@3}\tnote{b}}
  & \multirow{2}{*}{\textbf{Lat. (s)}} \\
\cmidrule(lr){2-3}\cmidrule(lr){4-6}
  & Total & Active & All & 95\% CI & Growth & & \\
\midrule
Qwen3.5~\cite{qwen2026next} & 0.9 & 0.9 & 0.14 & 0.07--0.21 & 0.04 & 0.11 & 6.6 \\
Qwen3.5 & 2 & 2 & 0.51 & 0.39--0.63 & 0.07 & 0.53 & 14.6 \\
Qwen3.5 & 4 & 4 & 0.72 & 0.61--0.84 & 0.30 & 0.54 & 11.1 \\
Qwen3.5 & 9 & 9 & 0.95 & 0.91--0.99 & 0.89 & 0.63 & 26.8 \\
Qwen3.5 & 27 & 27 & \textbf{0.99} & 0.97--1.00 & \textbf{0.96} & \textbf{0.67} & 26.2 \\
Qwen3.5 & 35 & 3 & 0.94 & 0.88--0.98 & 0.81 & 0.64 & 10.9 \\
\addlinespace[2pt]
Qwen3~\cite{yang2025qwen3} & 4 & 4 & 0.65 & 0.55--0.75 & 0.30 & 0.54 & 30.5 \\
Qwen3.8~\cite{qwen2026next} & 27 & 27 & 0.90 & 0.82--0.96 & 0.63 & \textbf{0.67} & 14.4 \\
\addlinespace[2pt]
Gemma4~\cite{gemma2026gemma4} & 8 & 4 & 0.90 & 0.84--0.94 & 0.89 & 0.62 & 15.3 \\
Gemma4 & 12 & 12 & 0.95 & 0.92--0.99 & 0.89 & 0.64 & 12.1 \\
Gemma4 & 26 & 3.8 & 0.96 & 0.91--1.00 & 0.93 & 0.65 & 8.2 \\
\addlinespace[2pt]
GPT-OSS~\cite{openai2025gptoss} & 21 & 3.6 & 0.52 & 0.40--0.65 & 0.22 & 0.41 & 11.4 \\
Granite4.2 & 8 & 8 & 0.81 & 0.69--0.91 & 0.70 & 0.57 & 47.1 \\
LFM2.5~\cite{liquid2025lfm2} & 8 & 1.5 & 0.27 & 0.16--0.39 & 0.07 & 0.19 & 24.2 \\
Muse Glimmer & 30 & 30 & 0.97 & 0.93--1.00 & 0.89 & \textbf{0.67} & 50.0 \\
\bottomrule
\end{tabular*}
\begin{tablenotes}\footnotesize
  \item[a] Active is lower than Total for mixture-of-experts models and, as effective size, for the 8B Gemma4.
  \item[b] EX is execution accuracy over all four missions with its 95\% bootstrap interval and over the growth mission alone; R@3 is scored on the three area-based missions, capped by a perception ceiling of 0.67, and Lat. is inference time per run.
\end{tablenotes}
\end{threeparttable}
\end{table}

\section{Conclusion}
\label{sec:conclusion}

This article presented a dual-speed agentic framework that makes MOT in satellite video self-configuring on board, where labels are absent, the detector stays frozen, and an observation lasts only tens of seconds. Its key idea is that the routine choice of a tracker configuration needs neither a search nor a reasoning turn on board, because the experience of labelled tasks can be compiled offline and only the final arbitration depends on the sequence in view. \auspex realizes this fast instinct with a covering shortlist and a label-free pairwise ranker, and \corvus carries it as a verified skill within a cascade of reflex, skill, and agent layers, so that the configuration is committed 2.5 s into each observation while a language-model agent, the slow reason, carries out missions stated in natural language. On two benchmarks with three recent detectors, the selector attains the best balance of accuracy and labels among sixteen methods at no more than 1.25 tracking passes, and the runtime executes missions without spending a language-model token on the routine decision while hosting fifteen language models without altering any committed configuration, which attests to the extensibility and generalization of the framework.

The evidence rests on two datasets and seven tasks, one of which carries most of the shift, and adapts only the association stage, while goal-level autonomy at ECSS level E4 remains outside the claims. Future work will widen the benchmark of shifted tasks, extend skills to further stages of the onboard chain, and co-design processors for the burst-and-wait load that \corvus exposes, toward an observation loop that closes entirely on board.

\bibliographystyle{IEEEtran}
\bibliography{refs}

\end{document}